\documentclass[sigconf]{acmart}

\usepackage{times}
\usepackage{latexsym}
\usepackage{xspace}
\usepackage{amsmath}

\usepackage{amssymb}
\usepackage{mathtools}
\usepackage{hyperref}
\usepackage{cleveref}
\usepackage{tikz}
\usepackage{graphicx}
\usepackage{booktabs}
\usepackage{array}
\usepackage{tabularx}
\usepackage{caption}
\usepackage{xurl}

\usepackage[most]{tcolorbox}
\usepackage{listings}
\usepackage{enumitem}
\usepackage{caption}
\usepackage{fvextra}

\usepackage{fancyvrb} 
\usepackage{multirow}

\usepackage{listings}
\usepackage[most]{tcolorbox}
\tcbset{listing engine=listings}
\usepackage{xcolor}  

\makeatletter\let\zifour@default\@undefined\let\zifour@scaled\@undefined\makeatother\usepackage{inconsolata} 
\usepackage{upquote}     

\usetikzlibrary{arrows.meta, positioning, calc, backgrounds}
\crefformat{section}{\S#2#1#3}
\crefformat{subsection}{\S#2#1#3}
\crefformat{subsubsection}{\S#2#1#3}

\usepackage[T1]{fontenc}
\usepackage[utf8]{inputenc}
\usepackage{microtype}
\usepackage[table]{xcolor}

\usepackage{graphicx}
\usepackage{stfloats}
\usepackage{pifont}

\usepackage{siunitx}
\newcommand{\res}[2]{#1 \textsubscript{$\pm$ #2}}
\newcommand{\cmark}{\textcolor{green!60!black}{\ding{51}}}      
\newcommand{\xmark}{\textcolor{red!80!black}{\ding{55}}}        
\newcommand{\pmark}{\textcolor{orange!70!black}{\boldmath$\triangle$}}

\newcommand{\namesoftware}{\texttt{IPO-Toolkit}\xspace}
\newcommand{\namedata}{\texttt{IPO-Dataset}\xspace}

\definecolor{codebg}{HTML}{F2F2F2}
\definecolor{codekw}{HTML}{569CD6}
\definecolor{codestr}{HTML}{C06040}
\definecolor{codecm}{HTML}{6A9955}
\definecolor{codetx}{HTML}{111111}

\definecolor{ImgBinA}{HTML}{F1F6FB} 
\definecolor{ImgBinB}{HTML}{E1EEF8} 
\definecolor{ImgBinC}{HTML}{CFE3F2} 
\definecolor{ImgBinD}{HTML}{B3D6EC} 
\definecolor{ImgBinE}{HTML}{8FC2E1} 
\definecolor{ImgBinF}{HTML}{6BAED6} 

\definecolor{ChartRow}{HTML}{DDECF7}

\newcommand{\bina}[1]{\cellcolor{ImgBinA}#1}
\newcommand{\binb}[1]{\cellcolor{ImgBinB}#1}
\newcommand{\binc}[1]{\cellcolor{ImgBinC}#1}
\newcommand{\bind}[1]{\cellcolor{ImgBinD}#1}
\newcommand{\bine}[1]{\cellcolor{ImgBinE}#1}
\newcommand{\binf}[1]{\cellcolor{ImgBinF}#1}

\newcommand{\capA}[1]{\colorbox{ImgBinA}{\vphantom{H}\,#1\,}}
\newcommand{\capB}[1]{\colorbox{ImgBinB}{\vphantom{H}\,#1\,}}
\newcommand{\capC}[1]{\colorbox{ImgBinC}{\vphantom{H}\,#1\,}}
\newcommand{\capD}[1]{\colorbox{ImgBinD}{\vphantom{H}\,#1\,}}
\newcommand{\capE}[1]{\colorbox{ImgBinE}{\vphantom{H}\,#1\,}}
\newcommand{\capF}[1]{\colorbox{ImgBinF}{\vphantom{H}\,#1\,}}

\lstdefinelanguage{PythonVS}{language=Python, morekeywords={match,case,nonlocal}}

\AtBeginDocument{%
  }

\copyrightyear{2026}
\acmYear{2026}
\setcopyright{cc}
\setcctype{by}
\acmConference[KDD '26]{Proceedings of the 32nd ACM SIGKDD Conference on Knowledge Discovery and Data Mining V.2}{August 09--13, 2026}{Jeju Island, Republic of Korea}
\acmBooktitle{Proceedings of the 32nd ACM SIGKDD Conference on Knowledge Discovery and Data Mining V.2 (KDD '26), August 09--13, 2026, Jeju Island, Republic of Korea}
\acmDOI{10.1145/3770855.3817580}
\acmISBN{979-8-4007-2259-2/2026/08}

\begin{document}

\title{IPO-Mine: A Toolkit and Dataset for Section-Structured Analysis of Long, Multimodal IPO Documents}

\author{Michael Galarnyk}
\authornote{Equal first authors: \{mgalarnyk3, slohani7\}@gatech.edu}
\affiliation{
  \institution{Georgia Institute of Technology}
  \city{Atlanta}
  \state{GA}
  \country{USA}
}

\author{Siddharth Lohani}
\authornotemark[1]
\affiliation{
  \institution{Georgia Institute of Technology}
  \city{Atlanta}
  \state{GA}
  \country{USA}
}

\author{Vidhyakshaya Kannan}
\affiliation{
 \institution{Sai University}
 \city{Chennai}
  \state{Tamil Nadu}
  \country{India}
}

\author{Sagnik Nandi}
\affiliation{
 \institution{Georgia Institute of Technology}
 \city{Atlanta}
  \state{GA}
  \country{USA}
}

\author{Aman Patel}
\affiliation{
 \institution{Georgia Institute of Technology}
 \city{Atlanta}
  \state{GA}
  \country{USA}
}

\author{Liqin Ye}
\affiliation{
 \institution{Georgia Institute of Technology}
 \city{Atlanta}
  \state{GA}
  \country{USA}
}

\author{Arnav Hiray}
\affiliation{
 \institution{Georgia Institute of Technology}
 \city{Atlanta}
  \state{GA}
  \country{USA}
}

\author{Rutwik Routu}
\affiliation{
 \institution{Duke University}
 \city{Durham}
  \state{NC}
  \country{USA}
}

\author{Prasun Banerjee}
\affiliation{
 \institution{Georgia Institute of Technology}
 \city{Atlanta}
  \state{GA}
  \country{USA}
}

\author{Siddhartha Somani}
\affiliation{
 \institution{Georgia Institute of Technology}
 \city{Atlanta}
  \state{GA}
  \country{USA}
}

\author{Sudheer Chava}
\affiliation{%
 \institution{Georgia Institute of Technology}
 \city{Atlanta}
  \state{GA}
  \country{USA}
}

\renewcommand{\shortauthors}{Michael Galarnyk, Siddharth Lohani, et al.}

\begin{abstract}
An Initial Public Offering (IPO) filing is a document released when a private firm goes public, allowing individual (retail) investors to purchase its shares. These filings describe a firm’s business, financials, and risks and are long, multimodal documents with narrative text and images. Despite their importance to financial markets, there is no large-scale, standardized dataset or benchmark for studying IPO filings with modern language and multimodal models. These documents pose significant challenges: filings frequently exceed 500{,}000 tokens and lack consistent structural organization. We introduce the IPO-Toolkit, an open-source framework for downloading and parsing IPO filings into standardized section-structured text and extracted images. The toolkit segments filings, extracts embedded images, and produces structured outputs that enable large-scale, reproducible analysis workflows over long, multimodal documents. Using this infrastructure, we construct the IPO-Dataset, a large, section-structured, multimodal dataset covering more than 109{,}000 IPO filings and amendments from 1994 to 2026 and containing over 76{,}000 images. We establish structured evaluation tasks over extracted financial charts, including chart quality and misleadingness assessment. Our experiments show that state-of-the-art multimodal models often diverge from expert human judgments on these tasks, exposing alignment challenges in multimodal reasoning over long, real-world regulatory documents. Beyond benchmarking, the IPO-Dataset enables large-scale analysis of section-level textual variation and cross-industry differences in visual and textual disclosure practices. Our \href{https://github.com/gtfintechlab/IPO-Mine/}{\textcolor{blue}{code}}, 
\href{https://huggingface.co/collections/gtfintechlab/ipo-mine}{\textcolor{blue}{dataset}}, 
and \href{https://ipomine.siddharthlohani.dev/}{\textcolor{blue}{website}} are publicly available under CC-BY-4.0.
\end{abstract}



\maketitle

\section{Introduction}
\label{section:introduction}

To begin trading on public markets, a company must file an Initial Public Offering (IPO) registration with the U.S. Securities and Exchange Commission (SEC), disclosing detailed information about its business operations, financial condition, and risk exposure. These disclosures are submitted as S-1 statements for domestic issuers and F-1 statements for foreign issuers and are organized into sections such as Risk Factors, Prospectus Summary, and Legal Matters. IPO filings play a critical role in shaping investor expectations, regulatory oversight, and capital allocation, particularly as individual (retail) participation in IPO markets grows and increasingly influences demand, pricing, and share allocations \citep{bloomberg2025retailipo}. 

Despite their importance, IPO filings are difficult to analyze. They routinely span hundreds of thousands of tokens and integrate heterogeneous content, including text, tables, and figures. As a result, direct end-to-end processing with large language models (LLMs) is both computationally expensive \citep{chen2024longloraefficientfinetuninglongcontext} and empirically unreliable at these lengths, even when inputs fall within claimed context limits \citep{he2025looglev2llmsready,zhang-etal-2024-bench,du-etal-2025-context}. Beyond document length, IPO filings pose a structural challenge: filings vary widely in section ordering, labeling, and formatting across issuers and over time. Reliable section-level and multimodal analysis therefore requires methods that operate on long, heterogeneous documents. Prior work on financial text analysis has focused on standardized filings such as 10-K annual reports \citep{wang-etal-2025-surf}. As a result, existing tools target these filings and replicate proprietary platforms \citep{lonare2021edgar,loukas-etal-2021-edgar,10.1145/3701716.3715289}, but do not support reliable section-level extraction or multimodal processing for IPO filings.

To address these challenges, we introduce a toolkit and dataset for large-scale, section-aware, multimodal analysis of IPO filings. The suite consists of two core components: IPO-Toolkit, a processing framework for S-1 and F-1 filings, and IPO-Dataset (see Table~\ref{tab:ipo-dataset}), a structured multimodal dataset constructed using this toolkit. Together, they enable scalable analysis of textual and visual patterns across industries and over time. Our contributions are as follows:
\begin{itemize}
    \item \textbf{IPO-Toolkit}: An open-source Python framework for scalable processing of IPO filings (S-1/F-1), supporting both HTML and legacy ASCII formats and enabling structured and multimodal extraction from long regulatory documents.
    \item \textbf{IPO-Dataset}: A large-scale, multimodal dataset of IPO filings (1994–2026) with section-aligned text and classified images to support longitudinal and cross-industry analysis.
    \item \textbf{Benchmark Suite}: Human-annotated benchmarks for section extraction and image classification, providing standardized evaluation tasks for IPO understanding.
    \item \textbf{Disclosure Analysis}: An empirical study of textual and visual disclosure trends using IPO-Dataset, including quantitative evaluation of misleading visual features across industries.

\end{itemize}

\begin{table}[t]
    \centering
    \renewcommand{\arraystretch}{0.85} 
    \setlength{\tabcolsep}{6pt} 
    \caption{Overview of the \namedata.}
    \label{tab:ipo-dataset}
    \begin{tabular}{l c}
        \toprule
        \textbf{Metric} & \textbf{Value} \\
        \midrule
        Number of Initial Filings & 20,819 \\
        Number of IPO Filings & 109,690 \\
        \quad \textit{  S-1 Filings} & 18,813 \\
        \quad \textit{  F-1 Filings} & 2,006 \\
        \quad \textit{  Amendments } & 88,871 \\
        Number of Images & 76,104 \\
        Human Annotated Images & 2,400 \\
        Years Covered & 1994--2026 \\
        \bottomrule
    \end{tabular}
\end{table}

\section{Related Work}
\label{section:related-work}

\subsection{Benchmarking on Financial Text}
Financial NLP benchmarks focus on earnings calls, 10-K filings, and news articles. FinBERT \cite{araci2019finbert} was an early model for sentiment analysis, followed by FLANG \cite{shah2022fluemeetsflangbenchmarks}, BloombergGPT \cite{wu2023bloomberggpt}, and FinGPT \cite{yang2023fingpt}. Benchmarks range from dictionary-based sentiment analysis \citep{loughran2011liability} to question answering and multi-label datasets, including FiQA \citep{FiQA}, SubjECTive-QA \citep{pardawala2025subjectiveqameasuringsubjectivityearnings}, and World Central Banks \citep{shah2025wordsuniteworldunified}.

\subsection{Multimodality in Finance}
Multimodal signals in finance include earnings calls, financial news, and investor-facing media. Applications include vocal cues from earnings calls \cite{sawhney-etal-2020-voltage}, numerical data fused with news \cite{ang-lim-2022-guided}, headline--market alignment in FNSPID \citep{10.1145/3637528.3671629}, visual sentiment in investor media \citep{gu2025giffluencevisualapproachinvestor}, financial short-form video captioning \citep{sukhani2025fincaptopicalignedcaptionsshortform}, and influencer stock recommendations from video, transcript, and audio signals \citep{10.1145/3711896.3737417}.

\subsection{Analyzing Charts and Visual Elements}
Visual elements are central to financial communication but remain difficult for models to interpret reliably. Benchmarks such as SciHorizon \citep{10.1145/3711896.3737403} and SPIQA \citep{pramanick2025spiqadatasetmultimodalquestion} evaluate structured reasoning and question answering over figures, while \citet{10.1145/3711896.3737437} show that categorical figure labels can obscure what models actually learn. Misleading-visualization studies examine deceptive chart features from web-sourced or synthetic examples \citep{10.1109/TVCG.2024.3456333, chen-etal-2025-unmasking}, and REVEAL \citep{10.24963/ijcai.2025/1081} shows that visual misinformation becomes harder to detect in multi-turn settings.

\section{IPO-Dataset and Construction}
\label{section:ipo-dataset}

We introduce IPO-Dataset, a large-scale, section-aligned, multimodal corpus of IPO filings from 1994--2026. The dataset is constructed from public SEC EDGAR filings and is not subject to copyright protection \citep{usc105}, enabling redistribution for research use. Dataset construction is supported by a Python toolkit that deterministically extracts document structure, filing text, and images.\footnote{\url{https://pypi.org/project/ipo-mine/}}

\subsection{IPO-Toolkit}
\label{subsec:dataset-toolkit}

We implement a structure-first dataset construction pipeline as a modular Python package built on top of the SEC EDGAR API. The toolkit enables deterministic, structure-aware extraction of IPO text and images directly from EDGAR source files and supports both legacy ASCII and modern HTML formats used in IPO filings. Although EDGAR provides programmatic access to filings, it lacks IPO-specific functionality required for analysis. For example, filings are indexed by Central Index Keys (CIKs) rather than tickers, recent and historical submissions are distributed across index tables, and embedded images are not directly exposed. IPO-Toolkit extends EDGAR with additional capabilities needed for dataset construction, including resolving ticker-to-CIK mappings, retrieving IPO filings across indices, downloading source documents, and aligning extracted text and images with filing-level metadata. Rather than relying on full-document processing, the toolkit leverages issuer-provided document structure \citep{Giovannini_2025_ICCV}. Filing Table of Contents (TOCs) are parsed to identify section names and page numbers, enabling segmentation into standardized sections while preserving original document organization. Because TOC parsing is a critical dependency for section segmentation, the toolkit includes automated validation and regression tests to ensure consistent behavior across filing formats and software updates. This design supports scalable processing across decades of filings while maintaining structural fidelity to the original documents. Appendix~\ref{section:additional-background} compares IPO-Toolkit with existing tools for structured SEC document processing.

\subsection{IPO Text}
\label{subsec:ipo-dataset-text}

\begin{figure}[t]
    \centering
    \includegraphics[width=\columnwidth]{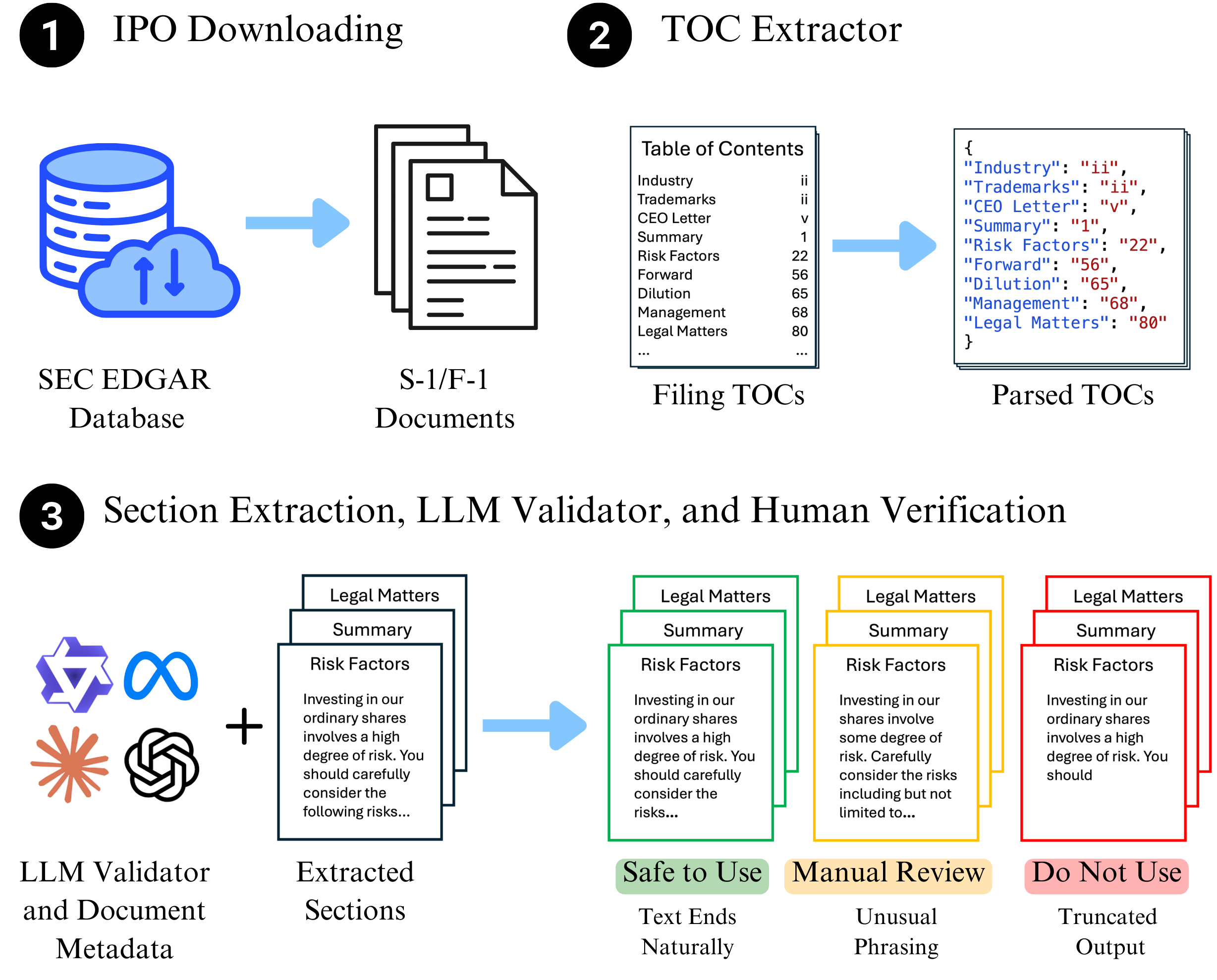}
\caption{
Overview of the IPO-Toolkit pipeline for constructing a section-structured IPO text corpus.
Filings are downloaded, TOCs are parsed, and text is extracted into standardized sections.
Automated checks with targeted human review label sections as \emph{Safe to Use}, \emph{Manual Review}, or \emph{Do Not Use}.
}
    \label{fig:software-pipeline}
\end{figure}

Figure~\ref{fig:software-pipeline} illustrates the three-stage workflow used to construct the section-structured IPO text dataset: (1) filing retrieval, (2) TOC parsing, and (3) section extraction with LLM-assisted validation and targeted human review. Rather than treating each IPO filing as a single block of text, the pipeline splits filings into standardized, section-level text that preserves document structure.

The resulting IPO text dataset contains standardized sections aligned to TOC-derived boundaries and normalized to a curated set of canonical section labels (e.g., Summary, Risk Factors, Legal Matters). Section names are normalized using fuzzy string matching against a curated dictionary of canonical labels \citep{hiray-etal-2024-cocohd}, ensuring consistency across issuers and years despite formatting variation. Each filing includes the original ASCII or HTML source, section-aligned text, associated images, and filing-level metadata such as accession number, filing date, and standard industry classification (SIC).

To ensure reliable section boundaries, we apply deterministic checks during TOC-based extraction to verify that sections are present, complete, and aligned with expected headers. This yields an average section extraction accuracy of 94.86\% across years. Because TOC-derived boundaries may produce truncated or misaligned sections, the pipeline incorporates a lightweight section-level validation step (Figure~\ref{fig:software-pipeline}, Stage 3). Extracted sections are labeled as \emph{Safe to Use}, \emph{Manual Review}, or \emph{Do Not Use} based on structural cues such as abrupt sentence endings or boundary spillover. The validator operates on surface-level completeness signals rather than semantic interpretation, enabling reliable quality control without full-document inference. This design avoids reliance on model-level financial knowledge that may exhibit temporal knowledge bias \citep{shah2025reportedcutofflargelanguage}.

\subsection{Verifying Text Section Extraction}
\label{section:verifying-section-extraction}

\begin{figure}[t]
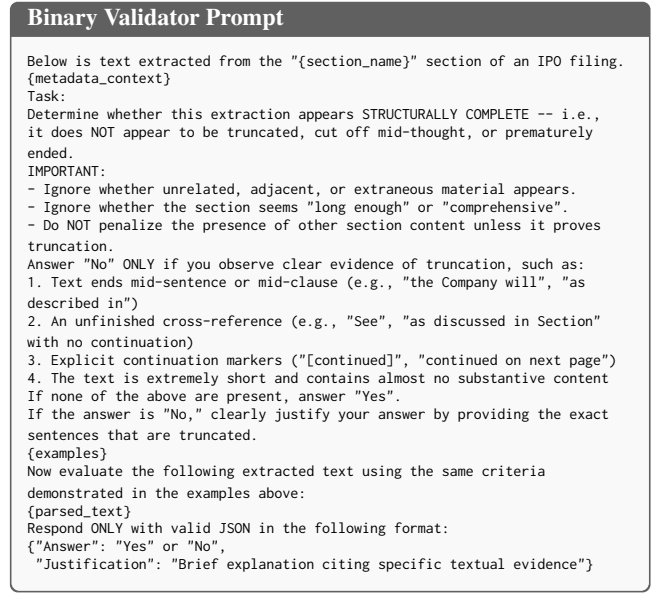

\centering

\begin{tcolorbox}[
  colback=gray!5!white,
  colframe=gray!75!black,
  title={\bfseries Binary Validator Prompt},
  width=\linewidth,
  boxrule=0.6pt,
  arc=3pt,
  outer arc=3pt,
  boxsep=2pt,
  left=4pt,
  right=4pt,
  top=4pt,
  bottom=4pt
]

\begin{Verbatim}[
  fontsize=\scriptsize,
  breaklines=true,
  breakanywhere=false,
  breaksymbolleft={},
  breaksymbolright={},
  obeytabs=false
]
Below is text extracted from the "{section_name}" section of an IPO filing.
{metadata_context}
Task:
Determine whether this extraction appears STRUCTURALLY COMPLETE -- i.e.,
it does NOT appear to be truncated, cut off mid-thought, or prematurely ended.
IMPORTANT:
- Ignore whether unrelated, adjacent, or extraneous material appears.
- Ignore whether the section seems "long enough" or "comprehensive".
- Do NOT penalize the presence of other section content unless it proves truncation.
Answer "No" ONLY if you observe clear evidence of truncation, such as:
1. Text ends mid-sentence or mid-clause (e.g., "the Company will", "as described in")
2. An unfinished cross-reference (e.g., "See", "as discussed in Section" with no continuation)
3. Explicit continuation markers ("[continued]", "continued on next page")
4. The text is extremely short and contains almost no substantive content
If none of the above are present, answer "Yes".
If the answer is "No," clearly justify your answer by providing the exact sentences that are truncated.
{examples}
Now evaluate the following extracted text using the same criteria demonstrated in the examples above:
{parsed_text}
Respond ONLY with valid JSON in the following format:
{"Answer": "Yes" or "No",
 "Justification": "Brief explanation citing specific textual evidence"}
\end{Verbatim}

\end{tcolorbox}

\caption{Binary validator prompt for IPO section completeness.}
\label{fig:validator_prompt}

\end{figure}

\begin{figure}[t]
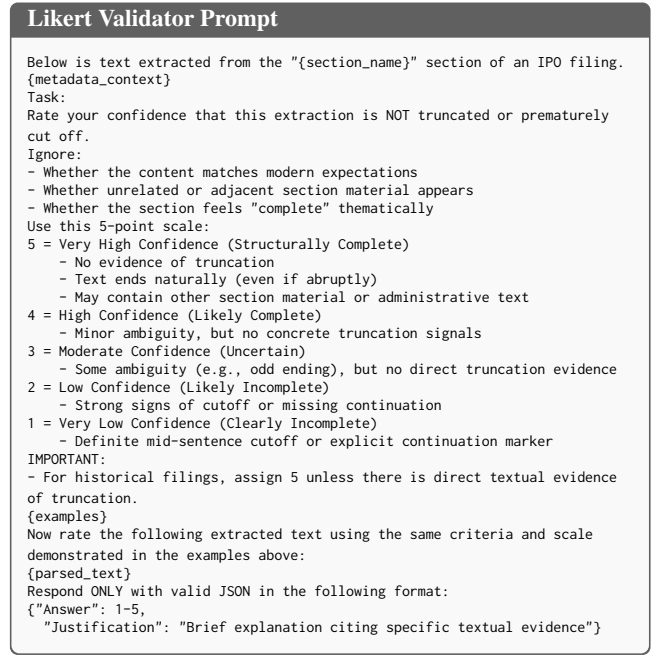

\centering

\begin{tcolorbox}[
  colback=gray!5!white,
  colframe=gray!75!black,
  title={\bfseries Likert Validator Prompt},
  width=\linewidth,
  boxrule=0.6pt,
  arc=3pt,
  outer arc=3pt,
  boxsep=2pt,
  left=4pt,
  right=4pt,
  top=4pt,
  bottom=4pt
]

\begin{Verbatim}[
  fontsize=\scriptsize,
  breaklines=true,
  breakanywhere=false,
  breaksymbolleft={},
  breaksymbolright={},
  obeytabs=false
]
Below is text extracted from the "{section_name}" section of an IPO filing.
{metadata_context}
Task:
Rate your confidence that this extraction is NOT truncated or prematurely cut off.
Ignore:
- Whether the content matches modern expectations
- Whether unrelated or adjacent section material appears
- Whether the section feels "complete" thematically
Use this 5-point scale:
5 = Very High Confidence (Structurally Complete)
    - No evidence of truncation
    - Text ends naturally (even if abruptly)
    - May contain other section material or administrative text
4 = High Confidence (Likely Complete)
    - Minor ambiguity, but no concrete truncation signals
3 = Moderate Confidence (Uncertain)
    - Some ambiguity (e.g., odd ending), but no direct truncation evidence
2 = Low Confidence (Likely Incomplete)
    - Strong signs of cutoff or missing continuation
1 = Very Low Confidence (Clearly Incomplete)
    - Definite mid-sentence cutoff or explicit continuation marker
IMPORTANT:
- For historical filings, assign 5 unless there is direct textual evidence of truncation.
{examples}
Now rate the following extracted text using the same criteria and scale demonstrated in the examples above:
{parsed_text}
Respond ONLY with valid JSON in the following format:
{"Answer": 1-5,
  "Justification": "Brief explanation citing specific textual evidence"}
\end{Verbatim}

\end{tcolorbox}

\caption{Likert validator prompt for IPO section completeness.}
\label{fig:likert_validator_prompt}

\end{figure}

We use a dual-validation prompt strategy combining a binary classification prompt (Figure \ref{fig:validator_prompt}) with a 1--5 Likert confidence score (Figure \ref{fig:likert_validator_prompt}). The binary output provides a clear filtering decision, while the Likert score captures graded confidence and helps identify borderline cases. Requiring agreement between both signals yields a conservative validation criterion, consistent with recent work emphasizing careful LLM-as-a-judge evaluation \citep{lee-etal-2025-checkeval}.
Sections are marked as \textit{SAFE TO USE} when the validator assigns a binary label of ``Yes'' and a Likert score $\geq 4$. Under this criterion, 22{,}666 documents (98.4\%) were automatically approved. The remaining 362 documents (1.6\%) were flagged for manual inspection before inclusion in the final dataset. To assess reliability, we manually review a subset of 200 sections labeled \textit{SAFE TO USE} and find that they are consistently free of truncation and boundary errors. This confirms that the dual-signal validator provides a high-precision filter for structurally valid sections.

\subsection{IPO Images}
\begin{figure}[t]
    \centering
    \includegraphics[width=\columnwidth]{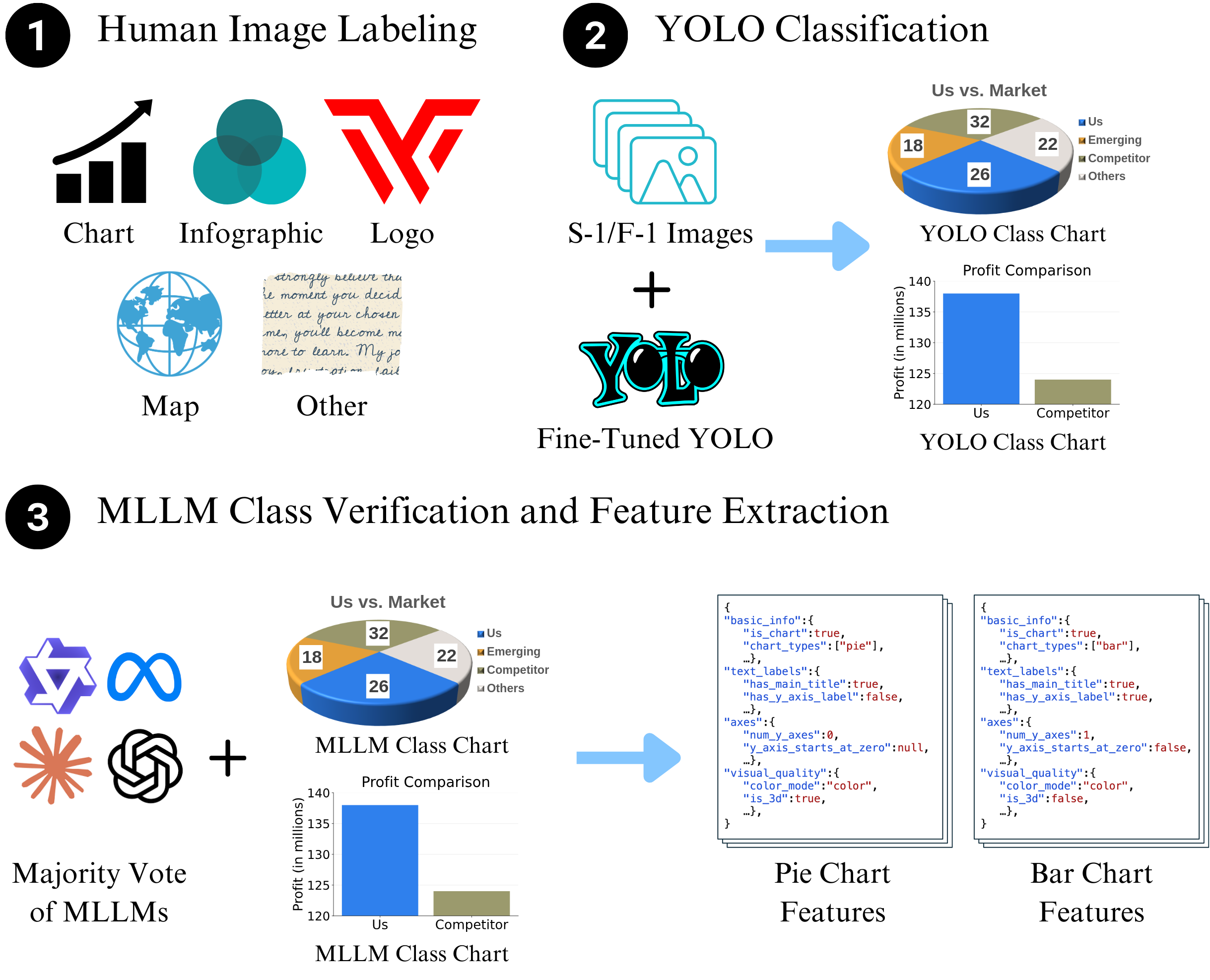}
    \caption{
    Image processing pipeline for IPO filings.
    Human-labeled images are used to fine-tune a YOLO model for large-scale image classification.
    Chart images identified by YOLO are verified and analyzed by MLLMs, with majority voting used to get structured chart features.
    }
    \label{fig:s1-image-pipeline}
\end{figure}

Figure~\ref{fig:s1-image-pipeline} illustrates the three-stage workflow used to construct the IPO image dataset: (1) human image labeling, (2) large-scale YOLO classification, and (3) class verification and feature extraction using multimodal large language models (MLLMs). The pipeline assigns each image a standardized class label and extracts structured visual attributes for charts.

The resulting dataset contains over 76{,}000 images extracted from IPO filings, including more than 17{,}000 charts. Each image is linked to its source filing and associated metadata, including accession number, filing date, and SIC. A comparison with existing vision-centric financial and chart benchmarks is provided in Appendix~\ref{section:additional-background}, where we contrast dataset scale, image composition, and document-level traceability. We define five mutually exclusive image classes: \textit{Charts, Infographics, Logos, Maps,} and \textit{Other}. The taxonomy was developed through an iterative pilot study in which approximately 1{,}000 images were manually labeled to refine category definitions and resolve ambiguous cases. The final label set emphasizes semantic intent (e.g., quantitative charts versus visually rich but non-quantitative infographics) rather than low-level visual cues. Each seed image is labeled by three annotators using a shared guideline, with majority voting used to resolve disagreements. Inter-annotator agreement statistics are reported in Table~\ref{tab:pairwise_krippendorff_ipo}.

To scale labeling across the full corpus, we fine-tune YOLOv8-small \citep{yolov8_ultralytics}, a widely adopted real-time vision model known for its balance of speed, accuracy, and computational efficiency \citep{talaat2023improved,xiao2024fruit,10.1145/3711896.3737395}. Using the human annotations, the model is trained to assign one of the five image classes at scale. Because distinctions between certain categories can be subtle, model predictions are subsequently verified using an ensemble of MLLMs. Labels are aggregated via majority voting, and images are retained when a majority of models agree on the assigned class. We analyze the sensitivity of this threshold and its relationship to human agreement in Table \ref{tab:image_agreement_thresholds}.

\begin{figure}[t]
\centering

\begin{tcolorbox}[
  colback=gray!5!white,
  colframe=gray!75!black,
  title={\bfseries Chart Classification Prompt},
  width=\linewidth,
  boxrule=0.6pt,
  arc=3pt,
  outer arc=3pt,
  boxsep=2pt,
  left=4pt,
  right=4pt,
  top=4pt,
  bottom=4pt
]

\begin{Verbatim}[
  fontsize=\scriptsize,
  breaklines=true,
  breakanywhere=false,
  breaksymbolleft={},
  breaksymbolright={},
  obeytabs=false
]
Analyze the provided image from an IPO filing to determine if it is a chart.
### System Constraints
* Output Format: Return VALID JSON only.
* No Markdown: Do NOT use code blocks (```json).
* No Commentary: Do not output any text before or after the JSON.
### Chart Definition
A chart is a quantitative visualization intended to convey numerical
trends, comparisons, or data patterns.
Charts include:
* Line charts showing trends over time
* Bar charts (vertical or horizontal)
* Pie charts or donut charts
* Scatter plots or bubble charts
* Area charts or stacked visualizations
* Histograms or distribution plots
* Choropleth maps or heat maps
* Other quantitative data visualizations
Non-charts include:
* Logos or brand marks
* Data tables or spreadsheets
* Maps or geographic visualizations
* Photographs or realistic images
* Signatures or handwritten text
* Letters or text documents
* Purely decorative graphics without data
### Classification Task
Determine whether the image is a chart based on the criteria above.
### Output JSON Schema
{"is_chart": boolean,
  "chart_justification": "string"}

Field Descriptions:
* is_chart: true if the image is a chart, false otherwise.
* chart_justification: Brief explanation (1-2 sentences) of why the
  image is or is not a chart.
\end{Verbatim}

\end{tcolorbox}

\caption{
Binary prompt for classifying chart images.}
\label{fig:is_chart_prompt}
\end{figure}

For images classified as charts, we extract structured visual attributes using a constrained MLLM prompt focused on observable visual properties. The feature schema captures four components: (1) basic metadata (e.g., chart type and presence of embedded data tables), (2) labels and legends (e.g., presence of descriptive axis labels and legends), (3) axis configuration (e.g., number of y-axes, whether the y-axis begins at zero, tick spacing consistency), and (4) visual quality indicators (e.g., use of 3D effects). Feature predictions are aggregated across multiple MLLMs via majority voting to reduce sensitivity to single-model errors. Detailed performance metrics and error analysis are reported in Appendix~\ref{app:model-error-analysis}.

\subsection{Human--YOLO Agreement}

Table~\ref{tab:pairwise_krippendorff_ipo} reports inter-annotator agreement on the human-labeled image subset used to fine-tune YOLO for large-scale classification. Each image was independently labeled by three annotators. Reliability was measured using Krippendorff's $\alpha$ \citep{Krippendorff2011ComputingKA}, which accounts for multiple raters and categories. Overall agreement is high ($\alpha = 0.911$). To examine disagreement patterns, we compute pairwise $\alpha$ for all unordered category pairs, restricting each calculation to images where annotators selected only the two categories under comparison. Disagreement is concentrated in comparisons involving \textit{Infographic}.

\begin{table}[!h]
\centering
\caption{
Pairwise inter-annotator agreement ($\alpha$) for the initial human-labeled image subset.
}
\label{tab:pairwise_krippendorff_ipo}

\renewcommand{\arraystretch}{0.85}
\setlength{\tabcolsep}{5pt}

\begin{tabular}{l c}
\toprule
\textbf{Category Pair} &
\textbf{Krippendorff’s $\alpha$ $\uparrow$} \\
\midrule
Chart vs.\ Logo & 1.000 \\
Chart vs.\ Map & 1.000 \\
Logo vs.\ Map & 1.000 \\
Chart vs.\ Other & 0.986 \\
Infographic vs.\ Logo & 0.980 \\
Chart vs.\ Infographic & 0.973 \\
Infographic vs.\ Map & 0.958 \\
Map vs.\ Other & 0.937 \\
Logo vs.\ Other & 0.908 \\
Infographic vs.\ Other & 0.793 \\
\midrule
\textbf{All categories (overall)} & \textbf{0.911} \\
\bottomrule
\end{tabular}
\end{table}

\subsection{MLLM--Human Agreement and Voting}

Table~\ref{tab:image_agreement_thresholds} reports how agreement among MLLMs relates to alignment with human annotations in the verification stage of the pipeline. Using outputs from 8 MLLMs, we group images by agreement level and measure alignment with human labels. For evaluation, we annotate all images with less than 100\% MLLM agreement and sample from the 100\% agreement bucket due to its larger size. Higher agreement among models consistently corresponds to stronger alignment with humans, indicating that inter-model agreement serves as a useful proxy for prediction confidence. Performance is relatively stable across agreement thresholds, suggesting that majority voting is a robust aggregation strategy. We therefore adopt a 50\% threshold as a simple and effective default for label aggregation.

\begin{table}[t]
\centering
\small
\caption{
Sensitivity of MLLM agreement thresholds for image classification.
Rows indicate the number of agreeing models (out of 8), and columns report predicted class counts.
For charts, we report MLLM--human match based on majority-vote labels.
}
\label{tab:image_agreement_thresholds}
\renewcommand{\arraystretch}{0.9}
\setlength{\tabcolsep}{2pt} 
\begin{tabular}{
l
c
@{\hspace{3pt}}!{\vrule width 0.7pt\hspace{3pt}}
c c c c
@{\hspace{3pt}}!{\vrule width 0.7pt\hspace{3pt}}
c
}
\toprule
& \multicolumn{1}{c@{}}{} &
\multicolumn{4}{c}{\textbf{Predicted}} &
\multicolumn{1}{c@{}}{\textbf{Chart Agreement}} \\
\cmidrule(lr){3-6}
\cmidrule(lr){7-7}
\textbf{Agree} &
\textbf{Total} &
\textbf{Info} &
\textbf{Logo} &
\textbf{Map} &
\textbf{Chart} &
\textbf{LLM--Human $\uparrow$} \\
\midrule
8/8 & 33,275 & 6,274 & 13,985 & 1,932 & 11,084 & 1,000/1,000 (100.00\%) \\
7/8 & 19,131 & 8,758 & 2,233 & 2,486 & 5,654 & 997/1,000 (99.70\%) \\
6/8 & 3,815  & 2,622 & 854 & 137 & 202 & 186/202 (92.08\%) \\
5/8 & 2,583  & 1,906 & 548 & 41 & 88 & 70/88 (79.55\%) \\
4/8 & 1,218  & 982 & 221 & 13 & 2 & 1/2 (50.00\%) \\
\bottomrule
\end{tabular}
\end{table}

\begin{table*}[t]
\centering
\small
\caption{
Yearly statistics for IPO filings by filing format and document length.
\textit{Format} denotes the SEC file type (ASCII/HTML).
The \textit{Firms} columns report the total number of IPO filers and the subset of firms containing at least one embedded image.
\textit{Images / Filing} reports the average number of embedded images and charts per filing.
Tokens/Filing and Tokens/Section columns report statistics computed using the
Llama-4 tokenizer~\cite{meta2024llama4}.
}
\label{tab:token_counts_by_year}

\renewcommand{\arraystretch}{0.85}
\setlength{\tabcolsep}{6pt}

\begin{tabular}{
l
c
!{\vrule}
c c
!{\vrule}
c c
!{\vrule}
c c
!{\vrule}
c
}
\toprule

& \multicolumn{1}{c@{}}{ } &
\multicolumn{2}{c}{\textbf{Firms}} &
\multicolumn{2}{c}{\textbf{Images / Filing}} &
\multicolumn{2}{c}{\textbf{Tokens / Filing}} &
\multicolumn{1}{c}{\textbf{Tokens / Section}} \\

\cmidrule(lr){3-4}
\cmidrule(lr){5-6}
\cmidrule(lr){7-8}
\cmidrule(lr){9-9}

\textbf{Year} & \textbf{Format} &
\textbf{Total} & \textbf{With Images} &
\textbf{Total} & \textbf{Charts} &
\textbf{ASCII/HTML} & \textbf{Filtered} &
\textbf{Risk Factors} \\
\midrule

1994 & \multirow{2}{*}{\begin{tabular}[c]{@{}c@{}}E-File\\Option\end{tabular}} &
35 & 0 (0\%) &
\res{0.0}{0.0} & \res{0.0}{0.0} &
\res{182,463}{115,884} & \res{182,463}{115,884} &
\res{3,646}{1,675} \\

1995 &  &
\cellcolor{gray!10}67 & \cellcolor{gray!10}0 (0\%) &
\cellcolor{gray!10}\res{0.0}{0.0} & \cellcolor{gray!10}\res{0.0}{0.0} &
\cellcolor{gray!10}\res{142,475}{75,123} & \cellcolor{gray!10}\res{142,475}{75,123} &
\cellcolor{gray!10}\res{4,300}{2,093} \\

\midrule
1996 & \multirow{4}{*}{\begin{tabular}[c]{@{}c@{}}ASCII\\Only\end{tabular}} &
708 & 0 (0\%) &
\res{0.0}{0.0} & \res{0.0}{0.0} &
\res{142,338}{97,567} & \res{142,338}{97,567} &
\res{5,238}{1,915} \\

1997 &  &
\cellcolor{gray!10}815 & \cellcolor{gray!10}0 (0\%) &
\cellcolor{gray!10}\res{0.0}{0.0} & \cellcolor{gray!10}\res{0.0}{0.0} &
\cellcolor{gray!10}\res{133,424}{105,533} & \cellcolor{gray!10}\res{133,424}{105,533} &
\cellcolor{gray!10}\res{5,321}{1,987} \\

1998 &  &
775 & 0 (0\%) &
\res{0.0}{0.0} & \res{0.0}{0.0} &
\res{151,146}{121,177} & \res{151,146}{121,177} &
\res{5,730}{2,015} \\

1999 &  &
\cellcolor{gray!10}707 & \cellcolor{gray!10}0 (0\%) &
\cellcolor{gray!10}\res{0.0}{0.0} & \cellcolor{gray!10}\res{0.0}{0.0} &
\cellcolor{gray!10}\res{129,447}{100,460} & \cellcolor{gray!10}\res{128,436}{98,534} &
\cellcolor{gray!10}\res{5,848}{2,062} \\

\midrule
2000 & \multirow{6}{*}{\begin{tabular}[c]{@{}c@{}}ASCII\\or\\HTML\end{tabular}} &
679 & 14 (2.1\%) &
\res{3.0}{2.5} & \res{0.1}{0.5} &
\res{140,889}{117,194} & \res{129,857}{109,702} &
\res{6,524}{1,952} \\

2001 &  &
\cellcolor{gray!10}255 & \cellcolor{gray!10}14 (5.5\%) &
\cellcolor{gray!10}\res{2.6}{1.3} & \cellcolor{gray!10}\res{0.4}{0.9} &
\cellcolor{gray!10}\res{116,446}{99,913} & \cellcolor{gray!10}\res{97,657}{60,003} &
\cellcolor{gray!10}\res{5,792}{2,329} \\

2002 &  &
537 & 103 (19.2\%) &
\res{3.2}{2.3} & \res{0.6}{1.3} &
\res{255,624}{277,639} & \res{138,387}{58,118} &
\res{6,064}{2,540} \\

2003 &  &
\cellcolor{gray!10}223 & \cellcolor{gray!10}103 (46.2\%) &
\cellcolor{gray!10}\res{5.3}{23.0} & \cellcolor{gray!10}\res{0.4}{0.9} &
\cellcolor{gray!10}\res{437,434}{531,847} & \cellcolor{gray!10}\res{119,250}{79,524} &
\cellcolor{gray!10}\res{6,761}{2,183} \\

2004 &  &
842 & 671 (79.7\%) &
\res{4.5}{5.2} & \res{0.5}{2.7} &
\res{758,999}{537,389} & \res{180,613}{90,335} &
\res{7,949}{2,116} \\

2005 &  &
\cellcolor{gray!10}444 & \cellcolor{gray!10}276 (62.2\%) &
\cellcolor{gray!10}\res{4.1}{4.6} & \cellcolor{gray!10}\res{0.8}{2.1} &
\cellcolor{gray!10}\res{495,387}{506,380} & \cellcolor{gray!10}\res{117,204}{81,279} &
\cellcolor{gray!10}\res{7,279}{2,362} \\

\midrule
2006 & \multirow{20}{*}{\begin{tabular}[c]{@{}c@{}}Mostly\\HTML\end{tabular}} &
708 & 558 (78.8\%) &
\res{3.0}{3.1} & \res{0.5}{1.8} &
\res{690,094}{520,644} & \res{175,834}{119,800} &
\res{7,384}{2,647} \\

2007 &  &
\cellcolor{gray!10}600 & \cellcolor{gray!10}420 (70.0\%) &
\cellcolor{gray!10}\res{4.5}{5.9} & \cellcolor{gray!10}\res{0.7}{2.0} &
\cellcolor{gray!10}\res{628,964}{527,778} & \cellcolor{gray!10}\res{138,530}{87,482} &
\cellcolor{gray!10}\res{8,086}{2,187} \\

2008 &  &
868 & 458 (52.8\%) &
\res{4.2}{5.2} & \res{0.8}{2.4} &
\res{507,818}{537,178} & \res{93,907}{77,197} &
\res{6,062}{2,892} \\

2009 &  &
\cellcolor{gray!10}719 & \cellcolor{gray!10}284 (39.5\%) &
\cellcolor{gray!10}\res{5.6}{14.1} & \cellcolor{gray!10}\res{2.4}{12.3} &
\cellcolor{gray!10}\res{540,824}{583,496} & \cellcolor{gray!10}\res{112,364}{103,323} &
\cellcolor{gray!10}\res{6,695}{2,750} \\

2010 &  &
852 & 501 (58.8\%) &
\res{4.5}{5.9} & \res{0.9}{2.9} &
\res{593,251}{543,764} & \res{113,632}{86,955} &
\res{6,542}{2,643} \\

2011 &  &
\cellcolor{gray!10}963 & \cellcolor{gray!10}655 (68.0\%) &
\cellcolor{gray!10}\res{4.0}{4.9} & \cellcolor{gray!10}\res{0.7}{2.2} &
\cellcolor{gray!10}\res{582,792}{471,728} & \cellcolor{gray!10}\res{115,774}{91,627} &
\cellcolor{gray!10}\res{6,861}{2,507} \\

2012 &  &
772 & 538 (69.7\%) &
\res{6.0}{4.8} & \res{0.8}{2.3} &
\res{803,347}{794,944} & \res{132,062}{110,615} &
\res{6,875}{2,727} \\

2013 &  &
\cellcolor{gray!10}678 & \cellcolor{gray!10}472 (69.6\%) &
\cellcolor{gray!10}\res{6.4}{12.6} & \cellcolor{gray!10}\res{2.0}{10.1} &
\cellcolor{gray!10}\res{567,364}{521,496} & \cellcolor{gray!10}\res{111,397}{84,409} &
\cellcolor{gray!10}\res{7,123}{2,662} \\

2014 &  &
886 & 512 (57.8\%) &
\res{7.5}{13.3} & \res{2.3}{10.3} &
\res{605,421}{626,288} & \res{114,793}{91,650} &
\res{7,054}{2,574} \\

2015 &  &
\cellcolor{gray!10}601 & \cellcolor{gray!10}448 (74.5\%) &
\cellcolor{gray!10}\res{6.6}{6.9} & \cellcolor{gray!10}\res{1.7}{3.5} &
\cellcolor{gray!10}\res{497,325}{481,589} & \cellcolor{gray!10}\res{103,440}{84,391} &
\cellcolor{gray!10}\res{7,173}{2,678} \\

2016 &  &
389 & 250 (64.3\%) &
\res{6.6}{8.9} & \res{1.7}{4.1} &
\res{480,232}{503,547} & \res{93,925}{83,328} &
\res{6,897}{2,721} \\

2017 &  &
\cellcolor{gray!10}438 & \cellcolor{gray!10}319 (72.8\%) &
\cellcolor{gray!10}\res{7.9}{11.4} & \cellcolor{gray!10}\res{1.8}{3.8} &
\cellcolor{gray!10}\res{521,253}{586,802} & \cellcolor{gray!10}\res{106,423}{91,388} &
\cellcolor{gray!10}\res{7,059}{2,926} \\

2018 &  &
479 & 345 (72.0\%) &
\res{9.9}{12.2} & \res{3.2}{6.0} &
\res{561,279}{663,917} & \res{108,833}{85,914} &
\res{7,217}{2,661} \\

2019 &  &
\cellcolor{gray!10}442 & \cellcolor{gray!10}348 (78.7\%) &
\cellcolor{gray!10}\res{10.5}{12.5} & \cellcolor{gray!10}\res{2.4}{5.0} &
\cellcolor{gray!10}\res{691,762}{720,346} & \cellcolor{gray!10}\res{133,956}{98,068} &
\cellcolor{gray!10}\res{7,217}{2,681} \\

2020 &  &
540 & 369 (68.3\%) &
\res{11.7}{14.2} & \res{3.1}{5.2} &
\res{648,644}{670,756} & \res{130,408}{86,279} &
\res{7,145}{2,903} \\

2021 &  &
\cellcolor{gray!10}1,621 & \cellcolor{gray!10}949 (58.5\%) &
\cellcolor{gray!10}\res{10.7}{13.0} & \cellcolor{gray!10}\res{2.5}{5.2} &
\cellcolor{gray!10}\res{746,067}{679,402} & \cellcolor{gray!10}\res{148,483}{75,960} &
\cellcolor{gray!10}\res{8,465}{1,717} \\

2022 &  &
609 & 452 (74.2\%) &
\res{7.3}{8.7} & \res{1.4}{3.2} &
\res{931,405}{878,375} & \res{151,227}{98,684} &
\res{7,672}{2,576} \\

2023 &  &
\cellcolor{gray!10}485 & \cellcolor{gray!10}379 (78.1\%) &
\cellcolor{gray!10}\res{8.9}{11.7} & \cellcolor{gray!10}\res{1.8}{4.2} &
\cellcolor{gray!10}\res{781,395}{934,000} & \cellcolor{gray!10}\res{125,559}{90,769} &
\cellcolor{gray!10}\res{7,350}{3,020} \\

2024 &  &
653 & 486 (74.4\%) &
\res{9.3}{12.6} & \res{2.1}{5.7} &
\res{725,949}{816,058} & \res{119,201}{84,795} &
\res{7,430}{2,900} \\

2025 &  &
\cellcolor{gray!10}1,223 & \cellcolor{gray!10}846 (69.2\%) &
\cellcolor{gray!10}\res{8.9}{11.1} & \cellcolor{gray!10}\res{1.7}{3.7} &
\cellcolor{gray!10}\res{738,071}{785,027} & \cellcolor{gray!10}\res{124,794}{86,966} &
\cellcolor{gray!10}\res{7,891}{2,640} \\

2026 &  &
206 & 139 (67.5\%) &
\res{10.8}{18.7} & \res{1.9}{5.7} &
\res{874,527}{835,035} & \res{137,670}{79,803} &
\res{7,651}{2,625} \\

\bottomrule
\end{tabular}
\end{table*}

\subsection{Filing Format Evolution and Dataset Implications}

Table~\ref{tab:token_counts_by_year} summarizes the evolution of IPO filing format, document length, and visual content over time. From 1994 to 1995, electronic submission was optional; between 1996 and 1999, filings were electronic but restricted to plain-text (ASCII), which does not support embedded images. In 2000, the SEC permitted HTML submissions and images \citep{sec_release_33_7684}, initiating a structural shift in how IPO filings were formatted. By 2006, the majority of IPO filings were submitted in HTML \citep{sec_edgar_filer_manual}. In financial disclosures, longer filings are associated with lower readability and greater difficulty in assimilating valuation-relevant information \citep{loughran2014measuring}.

The transition to HTML has two direct implications for dataset construction. First, images emerge and increase steadily following HTML adoption, consistent with the growth in images per filing reported in Table~\ref{tab:token_counts_by_year}. Second, raw token counts increase after HTML adoption due to markup and layout tags, while filtered text remains comparatively stable. Since filing length is often used as a proxy for readability and information-processing costs \citep{loughran2014measuring}, this raw-versus-filtered distinction matters for comparisons over time. The pattern suggests that much of the apparent growth in document length reflects formatting rather than substantive expansion of narrative disclosure. Consistent with this interpretation, we do not observe a clear increase in filtered IPO filing length around the Sarbanes--Oxley Act of 2002 \citep{sarbanes_oxley_2002}, unlike findings for 10-K filings \citep{10.5555/1620754.1620794,TSAI2017243}. Notably, filtered token counts for the \textit{Risk Factors} section remain relatively stable over time.

\section{Experiments}
\label{section:experiments}

\begin{table*}[t]
\centering
\small
\renewcommand{\arraystretch}{0.85}
\setlength{\tabcolsep}{3pt}

\caption{
Misleadingness scores (1 = highly accurate, 5 = highly misleading) for MLLMs on charts. \textbf{Industry Classification} columns report mean scores by industry, and \textbf{ALL} reports the overall mean.
\textbf{Visual Property Subsets} columns report mean scores restricted to charts exhibiting a given property:
3DP = 3D pie charts,
3DB = 3D bar charts,
2YA = charts with two y-axes,
TRC = truncated y-axis,
COM = combo charts,
GRA = grayscale charts,
COL = color charts.
Green cells denote the closest score to the human baseline.
}
\label{tab:industry_truthfulness_likert}

\begin{tabular}{
l
c c c c c c c c c c
!{\vrule width 0.7pt\hspace{3pt}}
c c c c c c c
}
\toprule

& \multicolumn{10}{c}{\textbf{Industry Classification}}
& \multicolumn{7}{c}{\textbf{Visual Property Subsets}} \\

\cmidrule(lr){2-11}
\cmidrule(lr){12-18}

\textbf{Model} &
\textbf{AFF} &
\textbf{MIN} &
\textbf{CON} &
\textbf{MAN} &
\textbf{TRN} &
\textbf{WHO} &
\textbf{RET} &
\textbf{FIRE} &
\textbf{SER} &
\textbf{ALL} &
\textbf{3DP} &
\textbf{3DB} &
\textbf{2YA} &
\textbf{TRC} &
\textbf{COM} &
\textbf{GRA} &
\textbf{COL} \\

\midrule

\rowcolor{gray!10}
\multicolumn{18}{c}{\textbf{Panel A: Average Scores}} \\
\midrule

\textbf{Human Baseline}
& 2.24 & 2.44 & 1.89 & 2.17 & 2.51 & 2.19 & 2.53 & 3.01 & 2.52 & 2.48
& 3.24 & 3.47 & 2.47 & 2.89 & 2.57 & 2.65 & 2.35 \\

\cmidrule(lr){1-18}

\textbf{MLLM Avg. (No CoT)}
& 2.49 & 2.68 & 2.60 & 2.68 & 2.62 & 2.38 & \cellcolor{green!20}2.63 & 2.70 & \cellcolor{green!20}2.62 & 2.61
& 2.48 & 3.13 & 3.00 & \cellcolor{green!20}2.85 & 2.92 & \cellcolor{green!20}2.62 & 2.61 \\

\textbf{MLLM Avg. (CoT)}
& \cellcolor{green!20}2.21 & \cellcolor{green!20}2.37 & \cellcolor{green!20}2.20 & \cellcolor{green!20}2.46 & \cellcolor{green!20}2.50 & \cellcolor{green!20}2.15 & 2.34 & \cellcolor{green!20}3.07 & 2.29 & \cellcolor{green!20}2.51
& \cellcolor{green!20}3.85 & \cellcolor{green!20}3.56 & \cellcolor{green!20}2.49 & 3.44 & \cellcolor{green!20}2.42 & 2.75 & \cellcolor{green!20}2.33 \\

\midrule

\rowcolor{gray!10}
\multicolumn{18}{c}{\textbf{Panel B: Model-Level Scores}} \\
\midrule

\multicolumn{18}{l}{\textbf{MLLMs (No CoT)}} \\[2pt]

Claude Opus 4.5
& 2.74 & 2.74 & 2.88 & 2.87 & 2.61 & 2.67 & 3.03 & 2.97 & 2.84 & 2.83
& 2.38 & 3.60 & 3.23 & \cellcolor{green!20}3.19 & 3.13 & 2.76 & 2.88 \\

Claude Sonnet 4.5
& 2.92 & 3.09 & 2.76 & 2.97 & 3.19 & 2.85 & 3.03 & \cellcolor{green!20}3.03 & 3.16 & 3.00
& 2.65 & 3.79 & 3.50 & 3.35 & 3.32 & 3.14 & 2.90 \\

GPT-5.2
& 2.87 & 3.01 & 2.95 & 3.29 & 3.04 & 2.63 & 3.05 & 3.35 & 2.99 & 3.08
& 3.60 & 3.83 & 3.17 & 3.36 & 3.22 & 3.11 & 3.05 \\

GPT-5-nano
& 2.03 & \cellcolor{green!20}2.44 & 2.35 & \cellcolor{green!20}2.21 & 2.20 & 1.82 & 2.06 & 2.09 & 2.07 & 2.13
& 1.72 & 2.33 & 2.75 & 2.35 & 2.69 & 2.10 & 2.16 \\

Qwen3-VL-32B
& 1.90 & 2.10 & 2.09 & 2.05 & 2.05 & 1.95 & 1.96 & 2.05 & 2.03 & 2.03
& 2.06 & 2.12 & 2.34 & 2.03 & 2.27 & 2.00 & 2.05 \\

\midrule

\multicolumn{18}{l}{\textbf{MLLMs (With CoT)}} \\[2pt]

Claude Opus 4.5
& 2.60 & 2.80 & 2.45 & 2.64 & 2.89 & 2.54 & \cellcolor{green!20}2.52 & 3.29 & 2.67 & 2.80
& 4.00 & 3.76 & 2.68 & 3.44 & \cellcolor{green!20}2.62 & 3.03 & 2.63 \\

Claude Sonnet 4.5
& \cellcolor{green!20}2.16 & 2.40 & \cellcolor{green!20}1.80 & 2.39 & \cellcolor{green!20}2.47 & \cellcolor{green!20}2.11 & 2.27 & 3.05 & 2.31 & \cellcolor{green!20}2.45
& 4.02 & \cellcolor{green!20}3.45 & 1.97 & \cellcolor{green!20}3.19 & 2.04 & 2.71 & \cellcolor{green!20}2.25 \\

GPT-5.2
& 2.44 & 2.47 & 2.23 & 2.62 & 2.56 & 2.41 & 2.59 & 3.18 & \cellcolor{green!20}2.57 & 2.67
& 3.97 & 3.59 & \cellcolor{green!20}2.46 & 3.45 & 2.47 & 2.85 & 2.53 \\

GPT-5-nano
& 2.11 & 2.22 & 2.33 & 2.45 & 2.37 & 1.97 & 2.17 & 2.78 & 1.97 & 2.36
& \cellcolor{green!20}3.36 & 3.21 & 3.19 & 3.35 & 2.88 & \cellcolor{green!20}2.63 & 2.17 \\

Qwen3-VL-32B
& 1.73 & 1.95 & 2.20 & \cellcolor{green!20}2.21 & 2.19 & 1.75 & 2.16 & \cellcolor{green!20}3.03 & 1.93 & 2.28
& 3.92 & 3.78 & 2.15 & 3.75 & 2.07 & 2.53 & 2.09 \\

\bottomrule
\end{tabular}
\end{table*}

\subsection{Detection of Misleading Financial Charts}

Misleading charts can distort interpretation through truncated axes, disproportionate scaling, dual y-axes, and 3D perspective effects. IPO filings provide a challenging setting because charts appear inside long regulatory documents, vary across industries, and combine standardized reporting with discretionary visual design. Because misleadingness is a subjective judgment task, we use expert human ratings as the reference point for evaluating model alignment \cite{galarnyk-etal-2025-inclusively}. We construct a stratified sample of 1,185 charts from IPO filings spanning nine industries. Each chart is labeled by three annotators on a 5-point Likert scale (1 = highly accurate, 5 = highly misleading), and we use the mean human rating as the baseline. We analyze results along two dimensions for human and model judgments: (i) industry aggregates (AFF, MIN, CON, MAN, TRN, WHO, RET, FIRE, SER) and (ii) visual-property subsets, including 3D pie charts (3DP), 3D bar charts (3DB), dual y-axes (2YA), truncated axes (TRC), combo charts (COM), grayscale charts (GRA), and color charts (COL). Models are evaluated under two prompting regimes:
(1) \textit{Direct prompting (No CoT)}, where the model assigns a score directly from the image; and
(2) \textit{Chain-of-thought prompting (CoT)}, where the model produces intermediate reasoning before outputting a final rating \cite{wei2023chainofthoughtpromptingelicitsreasoning}. In the CoT setting, we use a region-aware prompt that directs attention to potentially misleading regions (axes, scaling, legends, and perspective effects) before producing a misleadingness score \cite{chen-etal-2025-unmasking}.

\subsection{Models}

We evaluate the following MLLMs on IPO-Dataset images: Claude Sonnet 4.5, Claude Opus 4.5 \cite{anthropic2025claude}, GPT-5.2, GPT-5 Nano \cite{openai2025gpt5}, and Qwen3-VL-32B \cite{yang2025qwen3technicalreport1}. Implementation details are in Appendix \ref{sec:implementation-details}.

\subsection{Results}

Table~\ref{tab:industry_truthfulness_likert} reports average misleadingness scores on the 5-point scale. Columns under Industry Classification aggregate scores within each industry. Columns under Visual Property Subsets restrict evaluation to charts exhibiting a specific visual property (e.g., 3D pie charts, truncated y-axes, dual y-axes), pooled across industries.

\paragraph{Alignment under No CoT}
Under No CoT, the average MLLM score in the ALL industry column is close to the human baseline (2.61 vs. 2.48), but individual models differ substantially. GPT-5.2 and Claude Sonnet 4.5 assign higher misleadingness scores (3.08 and 3.00), while GPT-5-nano and Qwen3-VL-32B assign lower scores (2.13 and 2.03). This spread shows that alignment with expert judgments varies considerably across models.

\paragraph{Effect of CoT prompting}
CoT improves alignment in the ALL industry column, shifting the MLLM average from 2.61 to 2.51, closer to the human baseline of 2.48. This is consistent with misleading-chart benchmarks showing that reasoning guidance can improve alignment with human judgments \cite{chen-etal-2025-unmasking}.

\paragraph{Industry-level variation}
Industry differences are modest for most sectors. FIRE receives the highest human misleadingness score (3.01), and the CoT average remains similar (3.07). No CoT scores vary less across industries, assigning similar scores even to lower-scoring sectors such as CON and MAN. This suggests that CoT improves industry-level calibration.

\paragraph{Visual property subsets}
Visual-property subsets separate chart-design effects from industry composition. Humans assign the highest scores to 3D bar charts (3.47), 3D pie charts (3.24), and truncated-axis charts (2.89), while model scores also rise for several of these subsets.

\section{Text and Image Analysis}
\label{section:analysis}

\subsection{Lexical and Semantic Diversity}

We extract section text from IPO filings and apply LLM-based validation to remove truncated segments. The final corpus contains 29,032 validated sections, including Risk Factors (10,286), Prospectus Summary (7,346), and Legal Matters (5,396).

\begin{figure}[t]
    \centering
    \includegraphics[width=\columnwidth]{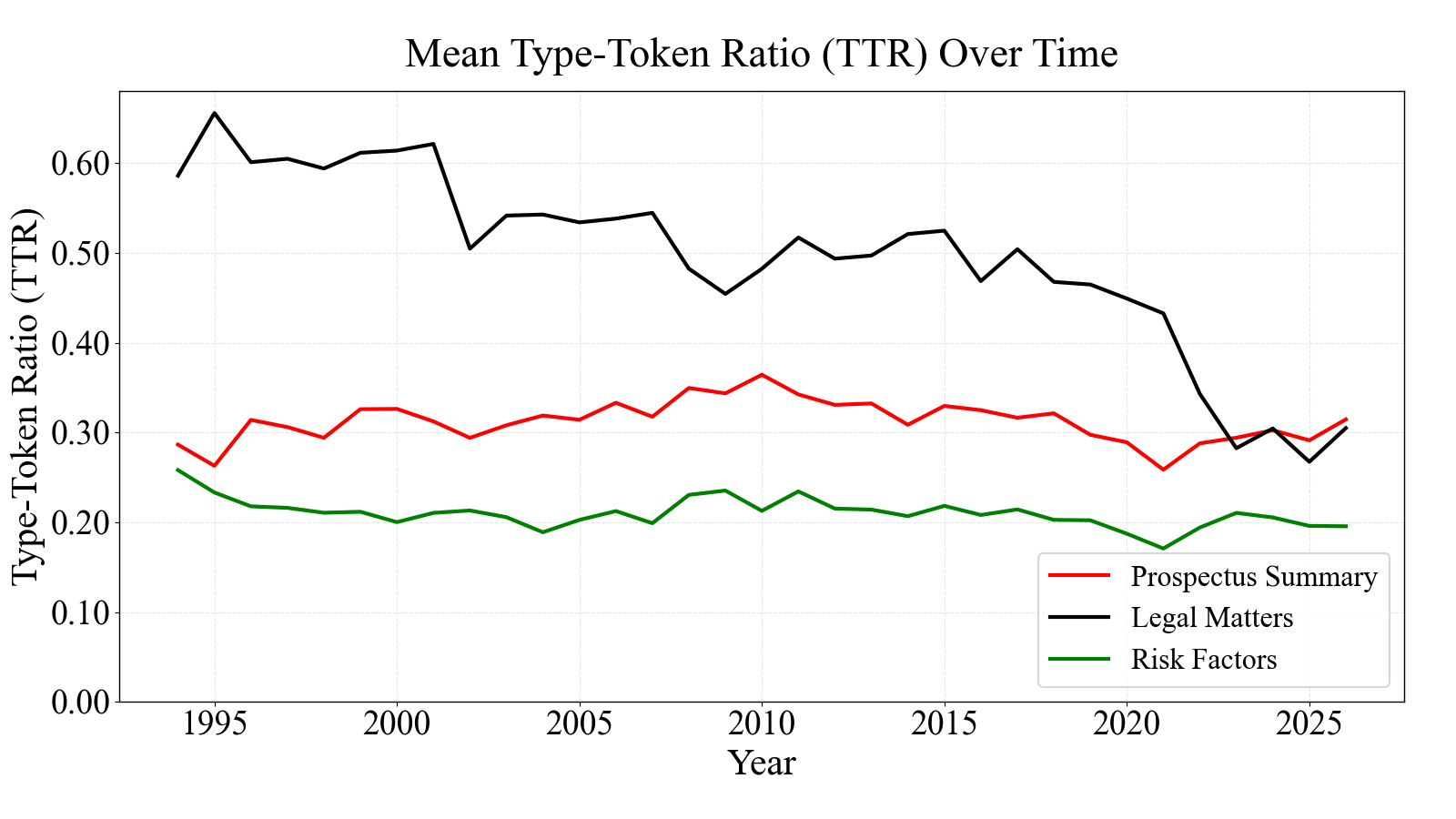}
\caption{
Mean TTR over time by disclosure section. Lexical
diversity is stable or declines across sections, with the steepest
reduction observed in Legal Matters, consistent with increasing
standardization and templated language in textual disclosures.}
    \label{fig:ttr_over_time}
\end{figure}

\paragraph{Lexical Diversity}
Figure~\ref{fig:ttr_over_time} reports lexical diversity over time using Type--Token Ratio (TTR) \cite{masry-etal-2025-chartqapro}. Because section lengths are comparable over time, TTR provides a consistent measure of lexical variation across years and sections. TTR is stable or gradually declining in all three sections, with the most pronounced reduction in Legal Matters. This pattern is consistent with increasing linguistic standardization in IPO filings, particularly in legally sensitive sections where templated language is common \cite{campbell2014risk, cohen2020resident}.

\begin{figure}[t]
    \centering
    \includegraphics[width=\columnwidth]{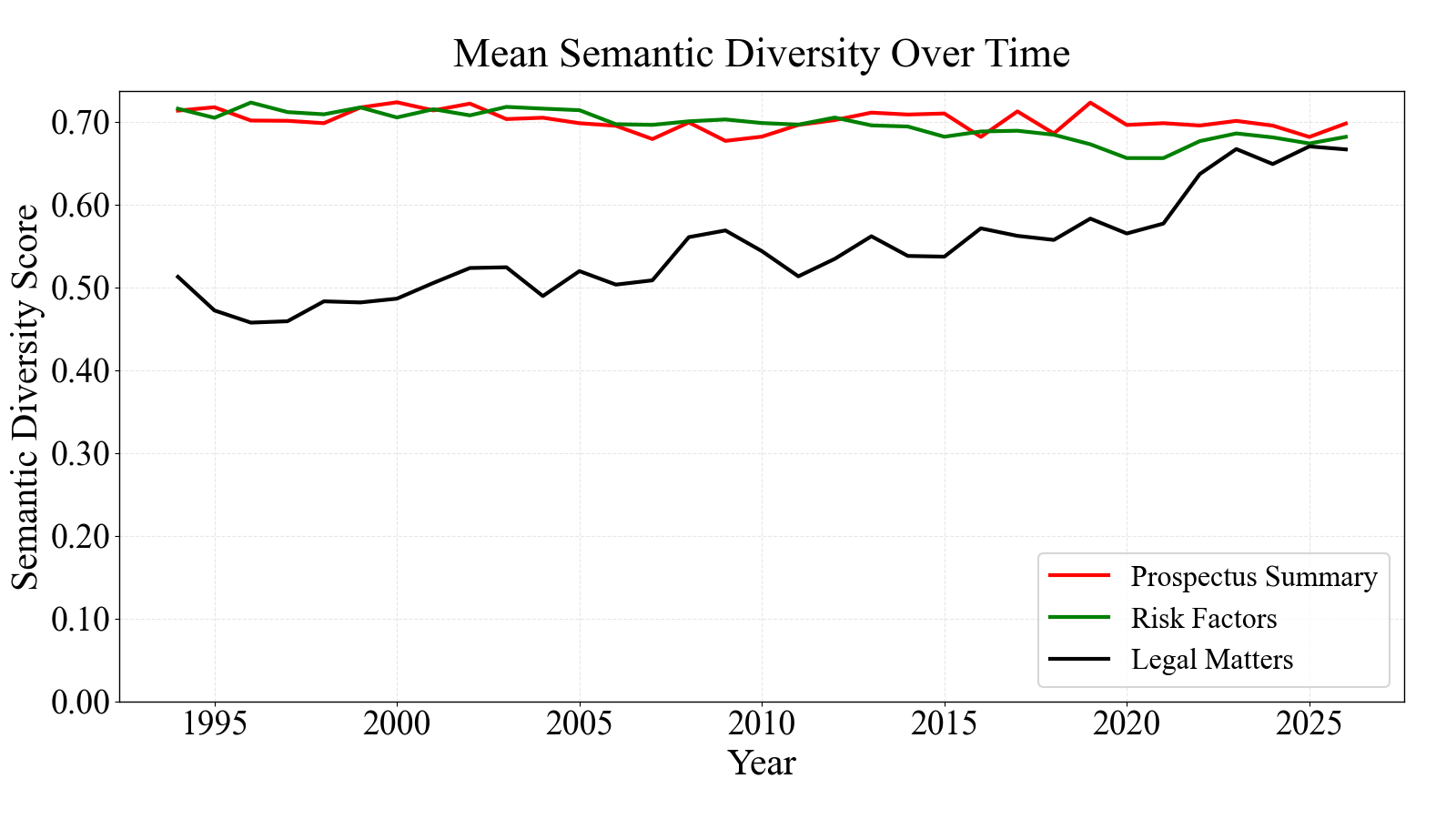}
\caption{
Mean semantic diversity over time by disclosure sec-
tion. The increase in semantic diversity in the Legal Matters
section is indicative of growing conceptual breadth in disclosures
despite increasing linguistic standardization.}
    \label{fig:semantic_over_time}
\end{figure}

\paragraph{Semantic Diversity}
Figure~\ref{fig:semantic_over_time} reports semantic diversity over time using sentence embeddings. For each section, we encode all sentences with a Sentence-BERT model \cite{reimers-gurevych-2019-sentence} and calculate the average pairwise cosine distance between embeddings. Higher average distance indicates greater conceptual spread within the section. Unlike TTR, semantic diversity increases over time, most notably in Legal Matters. This divergence suggests that while firms rely on increasingly standardized phrasing, they address a broader range of topics and contingencies. In other words, filings become more linguistically uniform yet conceptually more expansive, consistent with prior evidence on the evolution of corporate reporting \cite{dyer2017evolution, brown2011large}.

\subsection{Rising Visual Heterogeneity in IPO Filings}

\begin{table*}[t]
\centering
\small
\caption{
Image class distribution by industry (percentages within industry; counts in parentheses).
\textbf{Total} aggregates across industries.
Cells are colored by column-wise percentage bins
(\capA{<5\%}, \capB{5--10\%}, \capC{10--20\%},
\capD{20--30\%}, \capE{30--45\%}, \capF{>45\%}).
}
\label{tab:image_distribution_by_industry}

\renewcommand{\arraystretch}{0.85}
\setlength{\tabcolsep}{3pt}

\begin{tabular}{
l
c
c c c c c
}
\toprule

& &
\multicolumn{5}{c}{\textbf{Image Class Distribution}} \\

\cmidrule(lr){3-7}

\textbf{Industry} &
\textbf{Abbreviation} &
\textbf{Chart} &
\textbf{Infographic} &
\textbf{Logo} &
\textbf{Map} &
\textbf{Other} \\
\midrule

Agriculture, Forestry \& Fishing & AFF
& \binc{14.8\% (43)}
& \bind{26.4\% (77)}
& \bind{20.3\% (59)}
& \bina{4.1\% (12)}
& \bine{35.6\% (103)} \\

Construction & CON
& \binf{61.6\% (908)}
& \binc{10.5\% (154)}
& \binb{9.0\% (132)}
& \bina{2.6\% (38)}
& \binc{16.2\% (238)} \\

Finance, Insurance \& Real Estate & FIRE
& \bine{39.1\% (8{,}229)}
& \binc{19.8\% (4{,}169)}
& \binc{19.2\% (4{,}050)}
& \bina{4.2\% (889)}
& \binc{17.5\% (3{,}682)} \\

Manufacturing & MAN
& \bind{24.5\% (6{,}925)}
& \binc{19.7\% (5{,}571)}
& \binc{16.6\% (4{,}677)}
& \bina{1.1\% (319)}
& \bine{38.1\% (10{,}757)} \\

Mining & MIN
& \binc{17.6\% (856)}
& \bind{27.5\% (1{,}336)}
& \binc{16.6\% (808)}
& \bine{30.6\% (1{,}487)}
& \binb{7.8\% (380)} \\

Retail Trade & RET
& \binc{17.8\% (486)}
& \bind{28.2\% (770)}
& \binc{19.9\% (543)}
& \bina{2.2\% (60)}
& \bine{32.0\% (873)} \\

Services & SER
& \binc{14.6\% (2{,}384)}
& \bind{25.1\% (4{,}079)}
& \binc{19.2\% (3{,}119)}
& \bina{1.6\% (255)}
& \bine{39.3\% (6{,}398)} \\

Transportation \& Utilities & TRN
& \bine{30.6\% (1{,}814)}
& \binc{16.5\% (979)}
& \binc{19.1\% (1{,}132)}
& \bina{1.8\% (107)}
& \bine{32.0\% (1{,}896)} \\

Wholesale Trade & WHO
& \binc{19.1\% (265)}
& \binc{18.3\% (254)}
& \bind{25.8\% (357)}
& \binb{5.2\% (72)}
& \bine{31.6\% (438)} \\

Other/Unknown & OU
& \binc{12.4\% (776)}
& \bind{23.8\% (1{,}489)}
& \bine{38.6\% (2{,}421)}
& \bina{3.0\% (185)}
& \bind{22.6\% (1{,}373)} \\

\midrule
\textbf{Total} &
&
\bind{22.4\% (17{,}030)}
& \bind{20.6\% (15{,}657)}
& \binc{19.5\% (14{,}876)}
& \bina{4.0\% (3{,}014)}
& \bine{33.5\% (25{,}527)} \\

\bottomrule
\end{tabular}
\end{table*}

\begin{figure}[t]
    \centering
    \includegraphics[width=\columnwidth]{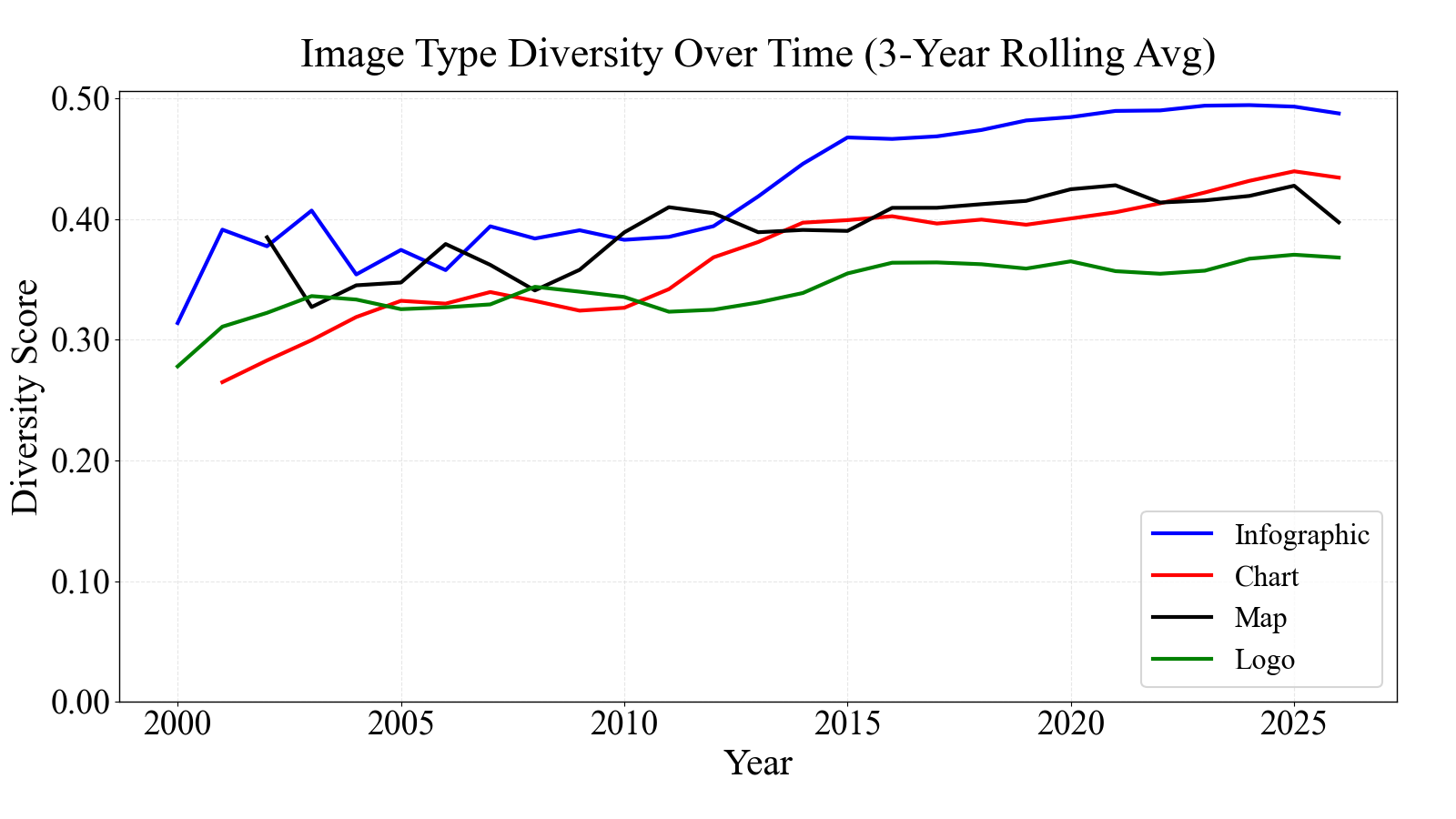}
    \caption{
    Longitudinal trends in image-type diversity (3-year rolling average). 
    Infographics (blue) and Charts (red) exhibit sustained increases in structural diversity.
    }
    \label{fig:diversity_over_time}
\end{figure}

The dataset contains over 76,000 images across five classes: charts, infographics, logos, maps, and other. Figure~\ref{fig:diversity_over_time} reports image-type diversity using a 3-year rolling average. In contrast to IPO text, IPO images have become more heterogeneous over time. To quantify visual diversity, we follow \citet{masry-etal-2025-chartqapro}, which encodes images into feature vectors using a CLIP vision encoder \citep{radford2021learningtransferablevisualmodels} and computes average pairwise cosine distances. By this measure, both Charts and Infographics become steadily more diverse over time.

This matters for model evaluation because visual diversity increases the difficulty of chart and figure understanding. Early chart benchmarks such as ChartQA \citep{masry-etal-2022-chartqa} relied on relatively constrained chart distributions, leading to performance saturation among MLLMs. ChartQA-Pro showed that model accuracy declines on more structurally diverse and visually complex charts \citep{masry-etal-2025-chartqapro}. The upward trajectory in IPO visual diversity suggests that financial filings increasingly resemble this more challenging regime rather than the simplified distributions in most benchmarks. These results highlight a growing modality gap: language becomes more templated over time, while visuals become more structurally diverse. Multimodal systems must therefore generalize across increasingly heterogeneous visual inputs embedded within standardized text.

\subsection{Industry-Level Variation in IPO Images}

Table~\ref{tab:image_distribution_by_industry} characterizes industry variation in IPO filing images, with the corresponding chart-class breakdown provided in Appendix~\ref{sec:chart_class_distribution}. Charts constitute 22.4\% of images, with prevalence varying by sector. Construction (CON) filings are the most chart-intensive (61.6\%), followed by Finance, Insurance \& Real Estate (FIRE) at 39.1\% and Transportation \& Utilities (TRN) at 30.6\%. Other sectors rely more heavily on non-chart visuals, including Agriculture (AFF), Services (SER), and Other/Unknown (OU). Across the dataset, the largest single category is \textit{Other} (33.5\%), encompassing photos, table renderings, text blocks, signatures, and product screenshots.

Certain visual types exhibit strong sectoral concentration. Maps are rare overall (4.0\%) but highly concentrated in Mining (MIN), where they account for 30.6\% of images, consistent with geographically grounded resource disclosures. Logos are especially prominent in Other/Unknown (OU) and Wholesale (WHO), reflecting branding-oriented filings. Manufacturing (MAN) and Services (SER) have the highest shares of \textit{Other} images, indicating greater use of photos, screenshots, text-heavy images, and other non-chart visuals.

Within charts, bar charts dominate (47.5\%), reflecting frequent categorical comparisons such as segment revenues or market composition. Line charts account for 22.9\% and are especially common in MAN and FIRE, where longitudinal trends are emphasized. Combination charts appear most often in CON and MIN, integrating levels and trends within a single visualization. Pie charts comprise just 10.1\%, indicating limited reliance on proportional encodings despite their common use in investor-facing materials.

\section{Conclusion}
\label{section:conclusion}

We present an open-source toolkit and dataset for section-structured, multimodal analysis of IPO filings. By combining issuer-provided document structure with automated processing and human validation, our work enables large-scale, reproducible extraction of textual sections and visual content from IPO filings spanning 1994–2026. Using the toolkit, we construct a dataset of over 109,000 filings containing more than 76,000 classified images. Our analysis reveals divergent trends in textual and visual disclosures: while text shows increasing standardization, visual content exhibits greater variation. These differences motivate benchmarks that evaluate visual design choices in realistic regulatory documents.

\clearpage
\newpage

\bibliographystyle{ACM-Reference-Format}
\bibliography{ref}

\newpage

\appendix
\clearpage

\relax

\section{Additional Background}
\label{section:additional-background}

We compare our work along two dimensions: SEC document processing infrastructure (Table~\ref{tab:tool_comparison_datasets}) and vision-centric image benchmarks (Table~\ref{tab:sota_finance_multimodal}). Existing tools primarily support text downloading and parsing for 10-K, 10-Q, and S-1/F-1 filings, but none integrate section extraction, image extraction, and LLM validation. IPO-Toolkit operates directly on IPO filings and combines these capabilities with open data and open code. Prior chart and financial image benchmarks rely on synthetic or scraped images and typically focus only on charts without document-level traceability. In contrast, IPO Images is constructed directly from SEC EDGAR filings (1994–2026), linking each image to its source filing and metadata, and spans charts, infographics, logos, and maps with human annotations at scale.

\begin{table*}[!tb]
\centering
\renewcommand{\arraystretch}{0.85}
\caption{
Comparison of tools for SEC document processing.
\textit{Modalities} lists supported data types: Text (T) and Image (I).
\textit{Open Data} and \textit{Open Code} indicate public availability.
\textit{Section Extractor}, \textit{Image Extractor}, and \textit{LLM Validation} denote supported capabilities.
}

\label{tab:tool_comparison_datasets}

\resizebox{0.85\textwidth}{!}{%
\begin{tabular}{lcccccccc}
\toprule
\textbf{Tool} & \textbf{Filing Type}
& \textbf{Modalities} & \textbf{Open Data} & \textbf{Open Code}
& \textbf{Section Extractor} & \textbf{Image Extractor}
& \textbf{LLM Validation} \\
\midrule

\textbf{sec-edgar-downloader} \citep{sec_edgar_downloader} &
10K/10Q, S-1/F-1 & [T] & \cmark & \cmark & \xmark & \xmark & \xmark \\

\textbf{sec-api.io} \citep{sec_api_io} &
10K/10Q, S-1/F-1 & [T] & \xmark & \xmark & \pmark & \xmark & \xmark \\

\textbf{EDGAR-Crawler} \citep{10.1145/3701716.3715289} &
10K/10Q & [T] & \cmark & \cmark & \cmark & \xmark & \xmark \\

\textbf{edgar (R package)} \citep{lonare2021edgar} &
10K/10Q, S-1/F-1 & [T] & \cmark & \cmark & \pmark & \xmark & \xmark \\

\midrule

\textbf{IPO-Toolkit (Ours)} &
S-1/F-1 & [T, I] & \cmark & \cmark & \cmark & \cmark & \cmark \\

\bottomrule
\end{tabular}
}
\end{table*}

\begin{table*}[!tb]
\footnotesize
\centering
\renewcommand{\arraystretch}{0.85}

\caption{
Comparison of vision-centric datasets for chart and image understanding.
Counts denote images per category.
\textit{Traceable} indicates whether images contain identifiers linking back to source documents.
\textit{Human} denotes the presence of human annotation.
}
\label{tab:sota_finance_multimodal}
\resizebox{0.85\textwidth}{!}{%
\begin{tabular}{lccccccccc}
\toprule
\textbf{Dataset} & \textbf{Years} & \textbf{Source} & \textbf{\# Images} & \textbf{\# Charts} & \textbf{\# Infographics} & \textbf{\# Logos} & \textbf{\# Maps} & \textbf{Traceable} & \textbf{Human} \\
\midrule

\textbf{ChartQA} \citep{masry-etal-2022-chartqa} &
2018-2022 & Statista+ &$\sim$21.9k  & $\sim$21.9k & 0 & 0 & 0 & \xmark & \pmark \\

\textbf{ChartQAPro} \citep{masry-etal-2025-chartqapro} & ---
 & Scraping/Tableau+ & $\sim$1.3k & $\sim$1.1k  & $\sim$190 & 0 & 0 & \xmark & \cmark\\

\textbf{InfographicVQA} \citep{9706887} &
--- & Scraping& $\sim$5.4k & 0 & $\sim$5.4k & 0 & 0 & \xmark & \cmark \\

\textbf{CharXiv} \citep{wang2024charxivchartinggapsrealistic} &
2020-2023 & arXiv & $\sim$2.3k & $\sim$2.3k & 0 & 0 & 0 & \cmark & \cmark \\

\textbf{MapQA} \citep{chang2022mapqadatasetquestionanswering} &
--- & KFF/Synthetic & $\sim$60k & 0 & 0 & 0 & $\sim$60k & \xmark & \xmark \\

\textbf{MME-Finance} \citep{10.1145/3746027.3758230} &
--- & Trading Platforms & $\sim$4.8k & $\sim$4.8k & 0 & 0 & 0 & \xmark &
\cmark  \\

\textbf{FinChart-Bench} \citep{shu2025finchartbenchbenchmarkingfinancialchart} &
2015-2024 & Bloomberg News+ & $\sim$1.2k & $\sim$1.2k & 0 & 0 & 0 & \xmark & \cmark  \\

\textbf{FinMME} \citep{luo-etal-2025-finmme} &
--- & Reports/Scraping & $\sim$4.4k & $\sim$4.4k & 0 & 0 & 0 & \pmark & \cmark \\

\textbf{ChartCheck} \citep{akhtar-etal-2024-chartcheck} &
--- & Wikipedia & $\sim$1.7k & $\sim$1.7k & 0 & 0 & 0 & \pmark & \cmark \\

\midrule
\textbf{\namesoftware} \textbf{(Ours)} &
\textbf{1994 - 2026} & SEC EDGAR & \textbf{$\sim$76k} & \textbf{$\sim$17k} & \textbf{$\sim$15k} & \textbf{$\sim$14k} & \textbf{$\sim$3k} & \textbf{\cmark} & \textbf{\cmark} \\

\bottomrule
\end{tabular}
}
\end{table*}

\section{Chart Class Distribution by Industry}
\label{sec:chart_class_distribution}

Table~\ref{tab:chart_type_distribution_by_industry} reports the distribution of chart classes within each industry. Percentages are computed within industry, with raw chart counts shown in parentheses.

\begin{table*}[t]
\centering
\caption{
Percentage distribution of chart classes within each industry.
Percentages are computed within industry, with raw chart counts shown in parentheses.
Each chart is assigned to one mutually exclusive class
(\textit{Bar}, \textit{Line}, \textit{Pie}, \textit{Combo}, or \textit{Other}).
The \textbf{Total} row aggregates counts across all industries.
Cells are colored using column-wise percentage bins
(\capA{<5\%}, \capB{5--10\%}, \capC{10--20\%},
\capD{20--30\%}, \capE{30--45\%}, \capF{>45\%})
to highlight differences in chart-class composition across industries.
}
\label{tab:chart_type_distribution_by_industry}

\renewcommand{\arraystretch}{0.85}
\setlength{\tabcolsep}{3.0pt}

\begin{tabular}{
l
p{1.9cm}<{\centering}
!{\vrule}
p{2.0cm}<{\centering}
p{2.0cm}<{\centering}
p{2.0cm}<{\centering}
p{2.0cm}<{\centering}
p{2.0cm}<{\centering}
}
\toprule

\multicolumn{2}{c@{}}{} &
\multicolumn{5}{c}{\textbf{Chart Class Distribution}} \\

\cmidrule(lr){3-7}

\textbf{Industry} &
\textbf{Abbreviation} &
\textbf{Bar} &
\textbf{Line} &
\textbf{Pie} &
\textbf{Combo} &
\textbf{Other} \\
\midrule

Agriculture, Forestry \& Fishing & AFF
& \binf{53.2\% (33)}
& \binc{19.4\% (12)}
& \binc{11.3\% (7)}
& \binb{9.7\% (6)}
& \binb{6.5\% (4)} \\

Construction & CON
& \binf{52.9\% (480)}
& \binc{17.3\% (157)}
& \bina{2.8\% (25)}
& \bind{26.7\% (242)}
& \bina{0.4\% (4)} \\

Finance, Insurance \& Real Estate & FIRE
& \binf{45.2\% (1{,}429)}
& \bind{22.9\% (726)}
& \binc{15.5\% (492)}
& \binc{14.5\% (459)}
& \bina{1.8\% (58)} \\

Manufacturing & MAN
& \binf{44.0\% (3{,}045)}
& \bine{30.1\% (2{,}081)}
& \binb{7.6\% (528)}
& \binc{13.3\% (920)}
& \binb{5.1\% (351)} \\

Mining & MIN
& \binf{41.0\% (281)}
& \binc{19.4\% (133)}
& \binb{9.1\% (62)}
& \bind{22.9\% (157)}
& \binb{7.6\% (52)} \\

Retail Trade & RET
& \binf{62.1\% (329)}
& \binb{7.5\% (40)}
& \binc{10.8\% (57)}
& \binc{18.1\% (96)}
& \bina{1.5\% (8)} \\

Services & SER
& \binf{60.4\% (1{,}439)}
& \binb{9.4\% (224)}
& \binb{8.8\% (209)}
& \binc{16.7\% (398)}
& \bina{4.8\% (114)} \\

Transportation \& Utilities & TRN
& \binf{45.8\% (609)}
& \bind{21.4\% (285)}
& \binc{11.8\% (157)}
& \binc{16.5\% (219)}
& \bina{4.6\% (61)} \\

Wholesale Trade & WHO
& \binf{58.5\% (155)}
& \binb{9.8\% (26)}
& \binc{18.1\% (48)}
& \binc{12.1\% (32)}
& \bina{1.5\% (4)} \\

Other/Unknown & OU
& \bine{38.0\% (295)}
& \bind{28.7\% (223)}
& \binc{17.9\% (139)}
& \binc{12.2\% (95)}
& \bina{3.1\% (24)} \\

\midrule
\textbf{Total} &
& \binf{47.5\% (8{,}095)}
& \bind{22.9\% (3{,}907)}
& \binc{10.1\% (1{,}724)}
& \binc{15.4\% (2{,}624)}
& \bina{4.0\% (680)} \\

\bottomrule
\end{tabular}
\end{table*}

\section{Model Error Analysis}
\label{app:model-error-analysis}

We analyze failure modes across components of the image pipeline to better understand their limitations on IPO filings. We first examine classification errors from the fine-tuned YOLO model trained on human-labeled images. We then analyze errors made by MLLMs when interpreting chart content, focusing on cases where image type was correctly identified but semantic interpretation failed.

\subsection{YOLO Classification Errors}
\label{section:yolo-error-analysis}

We manually reviewed instances where MLLM verification disagreed with YOLO predictions and grouped recurring error patterns. Failures are primarily driven by reliance on surface-level visual cues rather than semantic intent.

\paragraph{Overweighting Secondary Elements.}
A recurring failure occurs when YOLO assigns a label based on a prominent visual element, even when that element is decorative or secondary within the image (e.g.,~\href{https://www.sec.gov/Archives/edgar/data/1114841/000095010900500012/p60604.jpg}{\textcolor{blue}{1}},~\href{https://www.sec.gov/Archives/edgar/data/1280998/000104746904017652/g954024.jpg}{\textcolor{blue}{2}},~\href{https://www.sec.gov/Archives/edgar/data/1516912/000162828018005901/prospectussummaryb7.jpg}{\textcolor{blue}{3}}). In these examples, the images were classified as \textit{Map} because the model focused on small globe or world-outline graphics, even though those elements did not determine the primary purpose of the image. The predicted label reflects attention to a recognizable subcomponent rather than the overall function of the visual.

\paragraph{Overweighting Layout Structure.}

In a subset of errors, YOLO assigns labels based on layout patterns rather than the function of the image. Images containing rows, columns, panels, or segmented blocks are sometimes classified into structured visual categories even when they do not encode quantitative information. For example, tables (which should be labeled \textit{Other}) were classified as \textit{Infographic} because their grid layout resembles designed summaries (e.g.,~\href{https://www.sec.gov/Archives/edgar/data/1367777/000114420410062816/pg64f.jpg}{\textcolor{blue}{1}}). Similarly, some image collages and multi-panel layouts were labeled as \textit{Infographic} due to their structured composition (e.g.,~\href{https://www.sec.gov/Archives/edgar/data/947577/000104746910009824/g720312.jpg}{\textcolor{blue}{2}}). Product screenshots with grid-like interfaces, which belong to the \textit{Other} class, were also misclassified as \textit{Infographic} (e.g.,~\href{https://www.sec.gov/Archives/edgar/data/1479426/000104746910003510/g47389.jpg}{\textcolor{blue}{3}}). Segmented brand imagery was occasionally mislabeled as \textit{Map} or \textit{Infographic} (e.g.,~\href{https://www.sec.gov/Archives/edgar/data/1137778/000095012302002111/z50542a2z50542a4.gif}{\textcolor{blue}{4}}). In these cases, the prediction reflects how the image is arranged rather than what it communicates.

\paragraph{Size-Based Logo Misclassification}

YOLO sometimes classifies small, isolated images as \textit{Logo} regardless of their actual content. Decorative graphics, icons, or cropped design elements are labeled as logos primarily due to their compact size and bounded framing (e.g.,~\href{https://www.sec.gov/Archives/edgar/data/1318885/000104746905027347/g848237.jpg}{\textcolor{blue}{1}}). In these cases, the prediction is driven by size and visual isolation rather than identifiable branding or textual markers.

\subsection{Multimodal LLM Error Analysis}
\label{section:mllm-error-analysis}

To validate the quality of our MLLM-based chart classification, we manually annotated 2,285 images predicted as charts by the MLLM ensemble. Human annotators confirmed 98.6\% of these predictions as valid charts, indicating high precision. We qualitatively reviewed the remaining false positives to understand recurring failure modes.

\paragraph{Embedded Charts in Composite Images.}

MLLMs occasionally labeled screenshots of web dashboards, mobile applications, or presentation slides as charts when small chart components were embedded within a larger interface (e.g.,~\href{https://www.sec.gov/Archives/edgar/data/2003292/000200329225000036/financialsataglance1ia.jpg}{\textcolor{blue}{1}},~\href{https://www.sec.gov/Archives/edgar/data/1493318/000121390025039451/tproduct_012.jpg}{\textcolor{blue}{2}}). Although a chart element is present, the image functions primarily as a composite interface rather than a standalone data visualization. In these cases, the presence of any chart-like component appears sufficient to trigger a chart label.

\paragraph{Infographic Misclassification.}

MLLMs also misclassified process-oriented or narrative infographics as charts when these visuals conveyed sequential steps, conceptual flows, or categorical information rather than quantitative relationships (e.g.,~\href{https://www.sec.gov/Archives/edgar/data/1254348/000149315223034850/emulate_003.jpg}{\textcolor{blue}{1}},~\href{https://www.sec.gov/Archives/edgar/data/1598674/000104746920000699/g940216.jpg}{\textcolor{blue}{2}},~\href{https://www.sec.gov/Archives/edgar/data/1649094/000119312520162659/g802328g45d24.jpg}{\textcolor{blue}{3}}, and~\href{https://www.sec.gov/Archives/edgar/data/1843762/000121390021066498/timage_009.jpg}{\textcolor{blue}{4}}). Despite structural similarities to charts, these graphics do not encode numerical axes or data-driven comparisons.

\paragraph{Additional Error Patterns.}

Less frequent errors involved misinterpretation of axes, legends, or units in low-resolution, or visually ambiguous images (e.g.,~\href{https://www.sec.gov/Archives/edgar/data/1144354/000104746905020923/g775862.jpg}{\textcolor{blue}{1}}, ~\href{https://www.sec.gov/Archives/edgar/data/941221/000104746914007784/g257420.jpg}{\textcolor{blue}{2}}). These cases highlight the sensitivity of multimodal models to visual quality and contextual ambiguity.

\subsection{Interactive Web Interface}
\label{sec:web_interface}

\begin{figure}[t]
    \centering
    \includegraphics[width=\columnwidth]{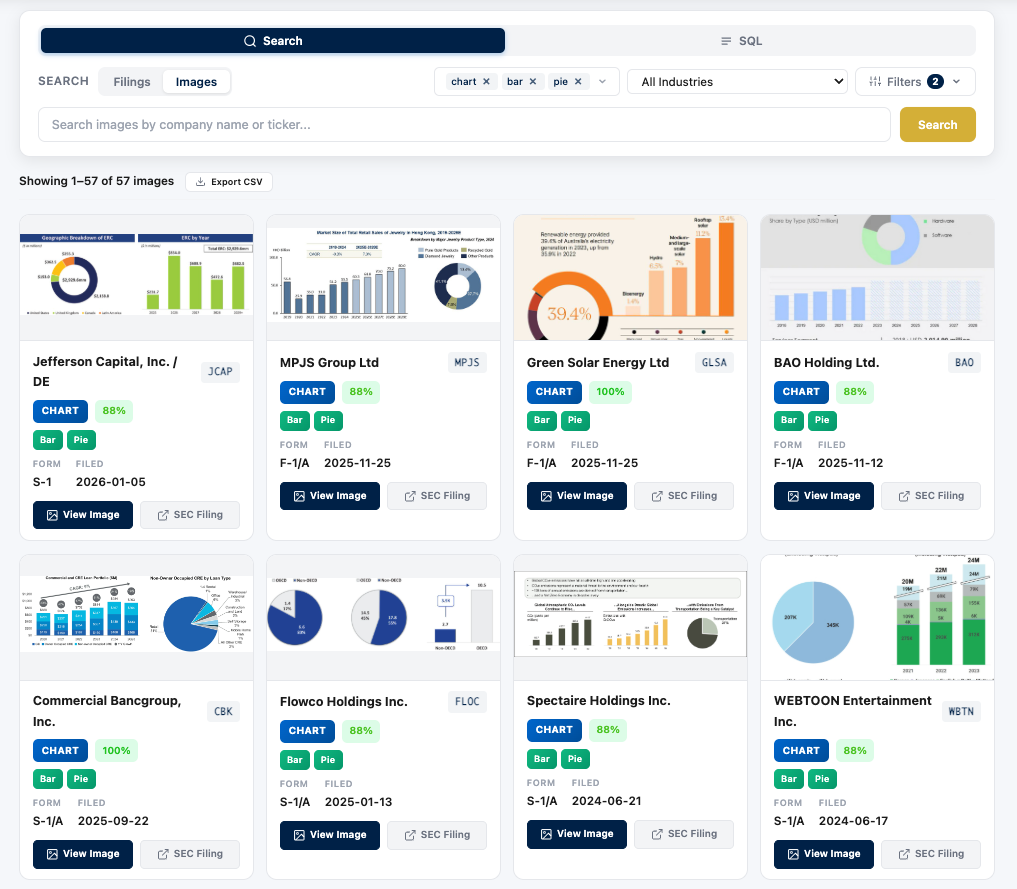}
    \caption{
    The web interface supports queries over companies, filings, and images using metadata and image attributes.
    }
\label{fig:ipo_mine_demo}
\end{figure}

Figure~\ref{fig:ipo_mine_demo} presents the IPO-Data web interface.\footnote{\url{https://demo.ipomine.siddharthlohani.dev}} In addition to the software package, we provide an open-source platform that enables interactive exploration of companies, filings, sections, and images. Users can filter results using standard metadata such as SIC industry, filing date, and form type, as well as structured attributes derived from the image classification and feature extraction pipeline, including image class, chart type, and prediction confidence thresholds.

The interface includes a built-in SQL console that allows custom queries against the dataset schema. This supports joins across filing metadata, text, and image-level attributes without local installation. The example shown retrieves images containing both bar and pie charts while excluding other visual types. Because filtering relies on structured attributes from the classification pipeline, users can target specific visual configurations and inspect results before export.

\section{Model Implementation Details}
\label{sec:implementation-details}

Model inference was conducted between January 5, 2026 and February 1, 2026. Anthropic models were accessed via the Anthropic\footnote{\url{https://github.com/anthropics/anthropic-sdk-python}} and GPT models via the
OpenAI API.\footnote{\url{https://openai.com/index/openai-api/}} Most open-source MLLMs were inferenced on an
NVIDIA RTX PRO 6000 Blackwell GPU.

\end{document}